\documentclass[runningheads,a4paper]{llncs}

\usepackage{amsmath}
\usepackage{amssymb}
\usepackage{times}
\usepackage{epsfig}
\usepackage{float}
\usepackage{mathtools}
\usepackage{booktabs}
\usepackage{array}
\usepackage[ruled,vlined,titlenumbered]{algorithm2e}
\usepackage{tikz}
\usepackage{wrapfig}

\usepackage{algorithmic}

\usepackage{xcolor}
\usepackage{soul}

\newcommand{\RR}{\ensuremath{\mathbb{R}}}

\newtheorem{df}{Definition}

\newtheorem{pte}{Property}

\usepackage{amsfonts, graphicx, amsopn, url, bm}
\usepackage{placeins}
\usepackage{verbatim}
\usepackage{color}

\definecolor{light-gray}{gray}{0.95}
\newcommand{\bb}{B\&B}
\newcommand{\bl}{\textit{BaGeL}}
\newcommand{\bg}{\textit{BaGeL}}

\newcommand{\normzero}[1]{{\|{#1}\|}_{0}}

\newcommand{\normtwo}[1]{{\|{#1}\|}_{2}}
\newcommand{\normP}[1]{{\|{#1}\|}_{p}}

\newcommand{\figsize}{5.9cm }

\begin{document}

\mainmatter  

\title{Constrained Machine Learning: The \bl{} Framework}

\author{Guillaume Perez$^1$, Sebastian Ament$^2$, Carla Gomes$^2$, Arnaud Lallouet$^3$}
\institute{
$^1$ Ugmented S.A.S, Sophia Antipolis, France\\
$^2$ Cornell University, Department of Computer Science, Ithaca NY 14850, USA\\
$^3$ Huawei Technologies Ltd, French Research Center\\
 \email{guillaume.perez06@gmail.com }}
\maketitle


\begin{abstract}
Machine learning models are widely used for real-world applications, such as document analysis and vision.  Constrained machine learning problems are problems where learned models have to both be accurate and respect constraints.  For continuous convex constraints, many works have been proposed, but learning under combinatorial constraints is still a hard problem.  The goal of this paper is to broaden the modeling capacity of constrained machine learning problems by incorporating existing work from combinatorial optimization.  We propose first a general framework called \bl{} (Branch, Generate and Learn) which applies Branch and Bound to constrained learning problems where a learning problem is generated and trained at each node until only valid models are obtained.  Because machine learning has specific requirements, we also propose an extended table constraint to split the space of hypotheses.  We validate the approach on two examples: a linear regression under configuration constraints and a non-negative matrix factorization with prior knowledge for latent semantics analysis. 

\end{abstract}

\section{Introduction}
The field of artificial intelligence has been drastically improved these last years, with the incredible results from image recognition and deep generative models \cite{he16,krizhevsky12,goodfellow14}.
In such applications an \textit{a priori} knowledge, such as, only a subset of inputs features are relevant (sparsity) \cite{candes07} may be present.
By design, the vanilla learning models are not well suited for learning under such assumptions.
For combinatorial constraints such as the bounded $l_0$ norm, which restricts the number of non-zero values, a huge research work has been done \cite{tibshirani96,donoho03}, and for some instances, the optimal model (i.e. the model respecting the constraint with the lowest loss function) can be found. 
But for the general case, it is still a hard question \cite{natarajan1995sparse}.
While many works have been done, it is most of the time an ad-hoc algorithm that heuristically enforces a given combinatorial constraint. For the sparsity, users will mostly use $l_1$ (i.e. sum of absolute values) relaxation instead of $l_0$ \cite{candes2008enhancing,slavakis10}.
Another example is on a recent work on topic modeling for materials science \cite{AIPhaseMapper}, that required that a set of combinatorial constraints on the solutions of a matrix factorization model to be satisfied. 
In such works, finding a machine learning model that does not respect the constraints is useless, even if the loss value is low.
Finding the prettiest relaxations, that are often both convex and differentiable is not necessarily useful, except if it allows to find a solution that at least does respect the constraints. 


On the other hand, Branch and Bound (\bb), is one of the most famous tools and is widely used to solve hard combinatorial optimization problems \cite{morrison16}. 
B\&B has been applied to many areas, and often, a dedicated version is extracted to fit the requirement of a given family of problems (branch and Price \cite{barnhart98}, branch and cut \cite{wolsey14} etc.).
B\&B is the main solving method of many combinatorial solvers, such as constraint programming solvers \cite{rossi06}, or even integer programming solvers \cite{wolsey14}.
These solvers have proven their efficiency in solving hard combinatorial problems.

The goal of this paper is to broaden the modeling capacity of constrained machine learning problems (CMLPs) by reusing the existing work of combinatorial optimization frameworks, and especially constraint programming.
In this paper, we introduce the \textit{Branch, Generate and Learn} (\bl).  This framework allows to learn while enforcing combinatorial constraints on the learned representation.  The goal of such a framework is to recursively generate learning problems in which the hypothesis space is restricted, until only valid hypothesis are allowed.
The difference with classic \bb{} framework is that at each node, a learning problem is generated and trained.
The generated problems are more and more restricted, until only valid models are obtainable.
The tree search part of the \bl{} ensures that all of the valid model space is explored.

Moreover, CMLPs have different needs than basic combinatorial problems.
That is the reason why a generalization of table constraints \cite{demeulenaere16,perez14} is proposed.
The table constraint is one of the most used constraints in constraint programming.
The proposed version is a hard, non-convex, and non-connected constraint, applied to a sub-sequence of the model's features.
This newly introduced constraint restricts the feature spaces into possibly non-connected and non-convex sub-spaces that act as bias on the hypothesis space.
Used with the \bl framework, combinations of this constraint will guide the search for valid models.
Moreover, it allows, for example, to fit data-sets using a prior knowledge, and generalizes many constraints such as the classic Lasso regularization constraint.

Finally, our experimental section shows on two hard combinatorial learning problems that the proposed method efficiently handles complex constraints and leads to good solutions.

\section{Modeling CMLPs}
A Constrained Machine Learning Problem (CMLP) is a learning problem in which restrictions are imposed on the hypothesis set, the learned representations, the inputs, or to any other parts of the learning process. 
A well known example is the Lasso regularization that add a $l_1$ regularization constraint to the learning model. 
In this example, the constraint is on the model parameters. 

\begin{df}
A Constrained Machine Learning Problem (CMLP) is a 5-uple $P=(V,M,C,f,DS)$ where $V$ is a set of variables, 
$M$ is a learning model and its parameter variables,
$C$ a set of constraints defined over a subset of the union of $V$ and $M$, 
$DS$ is a data set and $f$ a function of $V$, $M$, and $D_S$ to optimize. 
\end{df}
Each variable $v \in V$ is associated to a domain $D_v$ containing a finite set of values to which $v_i$ can be assigned.
The solution of a constrained learning problem is an assignment of variables $V$ and parameters of $M$ such that all the constraints in $C$ are satisfied and $f$ is minimized with respect to $DS$.


\begin{example} \label{ex:smart} \textbf{Smart Design.}
We consider the case of a Machine Learning model that can take as input multiple sensors.  Such component can be a camera (matrix of pixels), a radar or a lidar (point cloud), a mono-valued sensor such as temperature, pressure or contact.  The presence and the number of occurrences of these sensors will determine a cost for the system and an energy consumption.  The goal is to find an acceptable configuration of the system minimizing the function to be learned.  Using a $l_0$ regularization may suggest to use only some pixels of the camera, and some points of the radar, and therefore imposes both sensors to be present, maybe exceeding the maximum cost of the design.
\end{example}

For this smart design example, we consider optimizing the parameters of a linear regression model $M$ from a training set $D_{Train} = (X \in \mathbf{R}^{m*d}; y\in\mathbf{R}^{d})$.
Parameters of $M$ are components of $\theta=(\theta_1,\theta_2,...\theta_d)$.
The evaluation of the quality of a learning is given by the loss function $f=\normtwo{X\theta-y}$. 
Let $C=\{c_1,c_2,...,c_k\}$ be a set of $k$ components.
Each component can be a camera, a lidar, a multi-value sensors, etc. And each component is associated to a cost.
Each component $c_j = (i_j,w_j)$, where $i_j$ is an input tensor and $w_j$ a cost.
Let $B$ be the maximum allowed cost and $u=(u_1,...,u_k)$ be a Boolean vector representing the use of a given component. 
The following problem is the smart design problem:
\begin{equation}\label{eq:objsmart}
\underset{\theta \in \RR^d}{\text{minimize}} \normtwo{X\theta-y} 
\end{equation}
subject to
\begin{equation}\label{eq:smartActive}
    \forall i \in C, \forall j \in I_c, \normzero{\theta_j} \leq u_i
\end{equation}
\begin{equation}\label{eq:budget}
    \sum_{i}^{C} u_iw_i < B
\end{equation}
Where the $\normzero{v}$ is the total number of non-zeros values in a vector. 
In such settings, finding the optimal selection of $\theta_i$ such that the loss function is minimized and the budget is respected is a hard combinatorial problem. 
Note that when all the $w_i$ are equal to 1, and all components contain only 1 input, this is equivalent to the well known $l_0$ constraint,
a problem of utmost importance \cite{natarajan1995sparse,tibshirani96,donoho03,candes2008enhancing,slavakis10,bogdan2015slope,perez19}.

Consider a toy problem with $|C|=4$, 
$c_1$=$(\mathbf{R}^{64*64},10)$,
$c_2$=$(\mathbf{R}^{32*32},6)$,
$c_3$=$(\mathbf{R}^{32*32},5)$,
$c_4$=$(\mathbf{R}^{10},1)$. Let $B=12$. The possible solutions of equation (\ref{eq:budget}) are 
$u \in T_s = \{(0,0,0,0)$, $(0,0,0,1)$, $(0,0,1,0)$, $(0,0,1,1)$, $(0,1,0,0)$, $(0,1,0,1)$, $(0,1,1,0)$, $(1,0,0,0)$, $(1,0,0,1)\}$.
The optimal solution is the one optimizing equation (\ref{eq:objsmart}).

\subsection{Extended Table constraint}
\label{sec_tabl}
In addition to existing combinatorial constraints, machine learning specific needs lead to the definition of dedicated constraints.
Indeed, prior knowledge, sparsity, etc. are currently leading the research direction of the optimization methods for machine learning because of the uniqueness of the constraints.
This section introduces the extended table constraint, 
a generic constraint which will be used to split the features' search space into sub-spaces.

Table constraints, also named extensional constraints \cite{demeulenaere16,perez14}, are defined by the list of valid value assignments for a vector of discrete variables in constraint programming. 
They are widely used, mostly in industrial applications, because of their simplicity and expressiveness.
They are often built by extracting the solutions of sub-problems \cite{lhomme12,dekker17}. 
For example, consider the following binary table constraint enforcing that the variables must have a difference of $-2$ or $+1$.
For a pair of discrete variables $(a,b)$ associated with the domain $[0,4]$, the table $T$ is $((0,2),(1,3),(2,4),(1,0),(2,1),(3,2),(4,3))$.
Thus the assignment $a=0,b=2$ is valid since $a-b=-2$ and so is $a=3,b=2$.

We extend the standard table constraint definition for a given vector of variables $y \in \mathbb{R}^n$ and a set of n-tuples $T$, with a cost function $c:\mathbb{R}^n \times \mathbb{R}^n \rightarrow \mathbb{R}$, and a threshold value $a$.
\begin{df}
\label{def:tableConstraint}
A vector of variables $y$ is valid with respect to the
extended table 
constraint \textit{ET($y,T,c,a$)} if and only if there exists an n-tuple $t \in T$ such that $c(t, y) \leq a$.
\end{df}
Note that this definition is equivalent to classical table constraints if $a = 0$ and $c$ is the Euclidean distance function.
The cost function $c$ allows to represent the feasibility set and $a$ the distance of $y$ to it.
Extended table constraints aim at splitting the search space of the learned features into specific sub-spaces.

\begin{figure}[t]
\centering
\includegraphics[width=7.25cm]{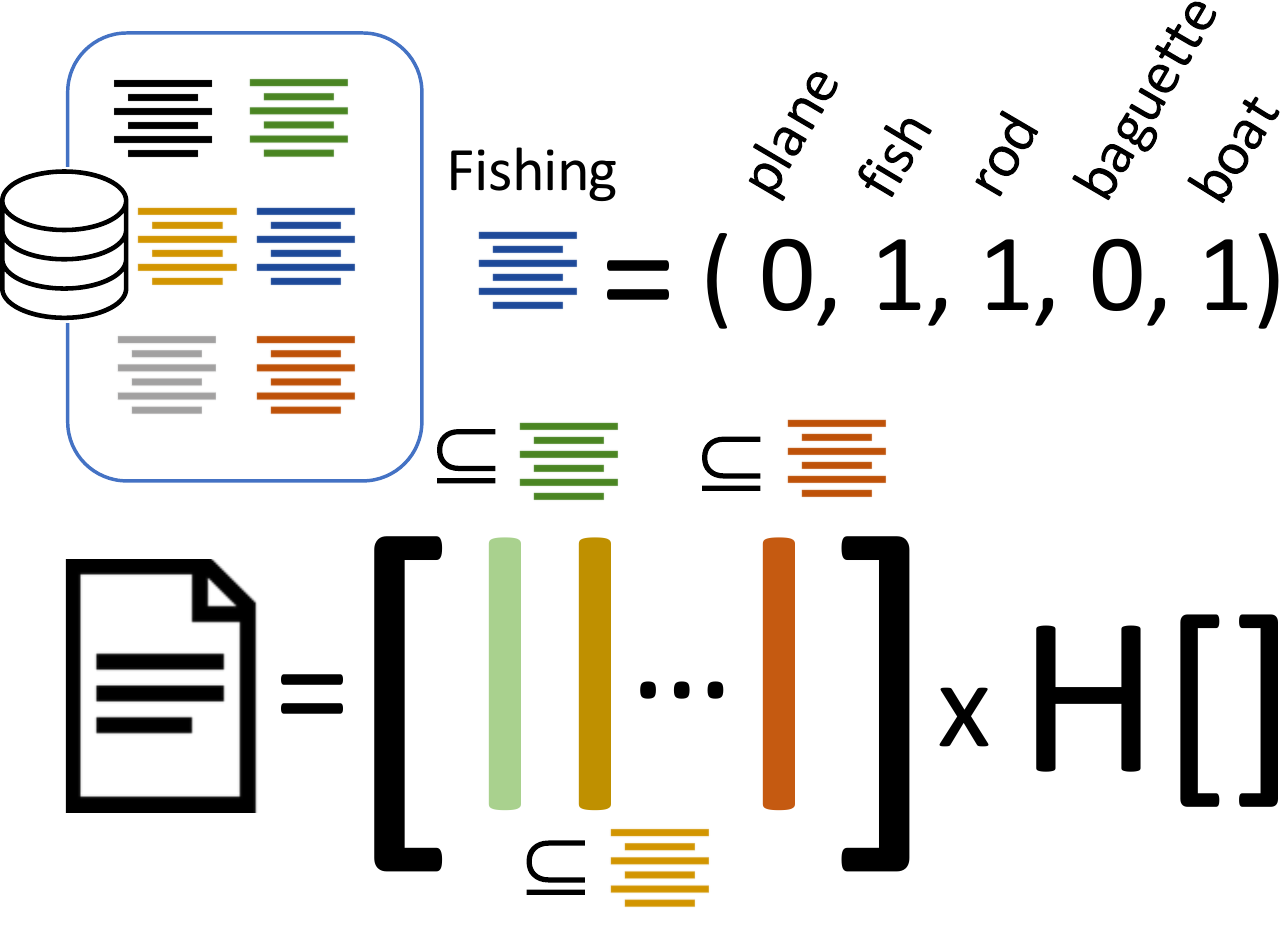}
\caption{Topic modeling problem with topics from a database. 
(top-left) A database of topics.
(top-right) An example of a topic vector: 1 indicates that the word belongs to the topic, 0 otherwise.
(bottom) Each column of the left matrix ($W$) is a subset of the words of a topic from the database.
}
    \label{fig:diagramTopic}
\end{figure}

\begin{example} \label{ex:NMF} \textbf{Prior NMF.}
To illustrate our concepts, we consider a version of a topic modeling problem from latent semantics analysis.  Extracting the topics of a text can be done using non-negative matrix factorization (NMF) \cite{NMFreview}. 
Given a matrix $A \in \RR^{n,m}$ representing a document, where each column represents the words of a paragraph,
we want to decompose $A $ into two non-negative matrices $W\in \RR^{n,k}$ and $H\in \RR^{k,m}$.  
The columns of $W$ represent the words that belong to a topic and $H$ represents the activation of the topics in each paragraph (or column) of $A$. 
The additional constraint is a restriction of the columns of $W$ that ensures that the topics are taken from an existing topic set.
\end{example}
Let $T$ be a table of prior topics, where each entry is a Boolean vector taking value \textit{1} if the word belongs to the topic and \textit{0} otherwise.
Let the function $c(y,t)=\normzero{y \cdot \overline{t}}$, with $\cdot$ being the Hadamard product.
Each column $w_i$ of matrix $W$ is constrained by an extended table constraint $C=ET(w_i,T,c,0)$.
Such constraint implies that the columns of matrix $W$ should match the database $T$.
Finally, a constraint ensures that all the selected topics from the prior data base are pairwise different.
An \textit{alldifferent} constraint \cite{regin1994filtering} is used.
Let $|T|$ be the number of prior topics.
Let $s_i \in [1,|T|]$ for $i \in [1,k]$ be decision variables representing the selected topic for column $i$.
$s_2=7$ implies that the second column of matrix $W$ is associated with topic $t_7 \in T$.
The problem can be defined as:
\begin{equation}
\underset{W,H}{\text{minimize}} \normtwo{A-WH} 
\end{equation}
subject to
\begin{equation}
    \forall i \in [1,k], ET(W_i, T, c, 0)
\end{equation}
\begin{equation}
    \forall i \in [1,k], s_i = j \implies c(w_i,t_j) = 0
\end{equation}
\begin{equation}
    AllDifferent(\{s_i| i \in [1,k]\})
\end{equation}

Figure \ref{fig:diagramTopic} illustrates this problem.
The function $c$ returns the minimum number of words to remove from a column of $W$ to match at least one topic from the database.
For example, with $w_1=(0.6,0.3,0.9,0,0)$ and the fishing topic $t_f$ from figure \ref{fig:diagramTopic}, we have $c(w_1,t_f)=\normzero{(0.6,0.3,0.9,0,0) \cdot (1,0,0,1,0)}=1$, which means there is one word that does not belong to the fishing vocabulary: \textit{plane}.
Let $a$ be the acceptable number of words that differentiate each vector in the solution from its nearest database entry.
If $a=0$, then no new word is accepted. Thus the constraint enforces that the columns of $W$ are a subset of the topics from the database. 

For discrete problems with finite domains, any constraint can be translated into a classic table constraint.
The only issue is that the size of this constraint may growth exponentially.
In the context of the extended table constraint, the distance function and threshold value help the representation of other constraints.
First, norm constraints such as $\normP{\theta} < \lambda$ \cite{natarajan1995sparse,tibshirani96,candes2008enhancing} can be translated into the extended table constraint $ET(\theta, \{(0,...,0)\},c(t,y)=\normP{y - t}, \lambda)$.
Moreover, the smart design constraints can be translated into an extended table too.
Consider the Example \ref{ex:smart} and its $u$ solutions table $T_s$.
Let $T_{se}$ be the set of tuples where for each tuple $t\in T_s$, there is a tuple $t_{se} \in T_se$.
For each value $t_j$ of $t$ associated to component $c_j=(i_j,w_j)$, a vector of the size of $i_j$ is created and filled with $t_j$.
Tuple $t_{se}$ is defined by the concatenation of these vectors.
Finally, constraint $ET(\theta,T_{se},c(y,t)=\normzero{y \cdot \overline{t}},0)$ is equivalent to constraint (\ref{eq:smartActive}) and (\ref{eq:budget}).

\section{\bg: A framework for CMLPs}
In this section the Branch, Generate and Learn (\bl) framework is presented. 
This framework is an adaptation of the \bb{} framework for CMLPs with combinatorial constraints. 
The main goal of \bl{} is to abstract the constraining of the learned representations from the learning process itself.
%
%
Analogously to \bb, \bl{} uses a tree search and needs to define the three main components, branching, search, and pruning \cite{morrison16}.
But the fourth and new most important component is the link between the learning process and the combinatorial process.
This link is the generation of learning problems in which the set of hypothesis is restricted.
These restrictions are going to be stronger and stronger, until only valid hypothesis are allowed.
%
%
The constrained generated problems take advantage of different methods, such as fixing some variables, modifying the configuration of the learning process, restricting the hypothesis space of the learning model using methods such as projections etc.
It should be composed of any existing machine learning problem, even another CMLP solved by \bl.

\noindent\textbf{Problem generation} Given the partial assignment $V^n$ of variables at node $n$, the generation function is $G(V^n,M^n) = P^n$, where $P^n$ is a learning problem. 
In addition, each solution of $P^n$ can be translated into a assignment of $M$.
Let $opt(P)$ be the optimal solution of problem $P$.
Let $Space(P)$ be the search space of problem $P$.
Let $\Theta_C$ be the space of all models satisfying the constraints.
\begin{df} \label{prop:leaf1}
A node $n$ is a leaf if $Space(P^n) \subseteq \Theta_C$.
\end{df}
This proposition implies that if the generated problem' solutions necessarily respect the constraint, there are no reason to restrict the space more.
Let $parent(n)$ be the parent nodes of node $n$.
\begin{df}
A \bl{} problem is monotonic restrictive if and only if
$$\forall n, \forall n' \in parent(n), Space(G(n)) \subseteq Space(G(n'))$$
\end{df}

\begin{pte}
Let a \bl{} minimization problem be monotonic restrictive.
Let $n^*$ be the best leaf (solution) found so far, and $n$ be the current node.
if $opt(n) \geq opt(n^*)$, then the node $n$ can be safely pruned.
\end{pte}
The proof is omitted as this property is classic for \bb{} like algorithms.
It allows \bl{} to stop exploring unpromising branches of the search tree.

The remaining case is when a node is a leaf with respect to definition \ref{prop:leaf1},
but finding the optimal solution is a hard problem.
For these cases, \bl{} proposes to sample the solution space to evaluate nodes.
In the general case, dedicated evaluation functions should be defined.

\begin{pte}
Let a \bl{} model be monotonic restrictive.
A node $n$ is a leaf if $opt(P^n)$ is found and $opt(P^n) \in \Theta_C$.
\end{pte}
If the optimal model of a node respects the constraint, it is unnecessary to search anymore the sub-tree emanating from this node.
This characteristic is unusual for \bb{} like algorithms, and allows to strongly cut the search space. 
Consider for example the smart design problem, an assignment of $\theta$ is enough to check if it is valid.
In the general case, an assignment of parameters of $M$ might not be enough to check the validity.
This is because variables of $V$ might need a deeper search for satisfiability.

A possible implementation of the \bl{} framework is given in Algorithm \ref{a:branchAndLearn}. 
The algorithm starts by opening the root node. 
Then the main loop of the algorithm will be run until no more nodes are open, or a stopping condition is reached. 
The stopping condition is implemented by the method $shouldStop()$.
Each time a node is picked by the search strategy, the \bl{} framework starts by enforcing the pruning rules, generates the learning problem and then trains it. 
Finally, if the node is not a leaf, decisions are generated by the branching strategy and added to the current set of open nodes. 
The goal of these decisions is to restrict more and more the search space, such that the deeper we are in the tree, the more constrained the learning problem will be.
Concretely, in each node, the pruning phase reduces the search space allocated to the learning model, and the learning phase tries to find the best solution with respect to these restrictions.

\begin{figure}
    \centering
    \includegraphics[width=8cm]{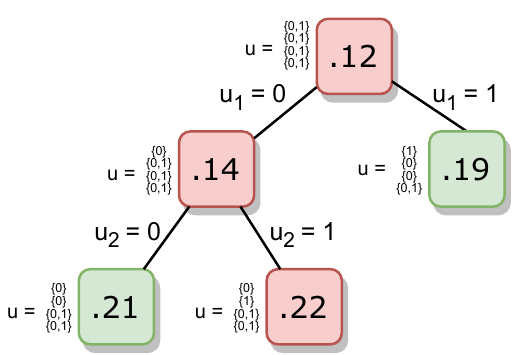}
    \caption{Branch and Generate algorithm for a Smart Design. Green node are solutions that respect the constraint, red nodes don't. The value inside the node is its loss.} 
    \label{fig:ex_123}
\end{figure}

Consider the Smart design problem instance defined in Example \ref{ex:smart}.
Let the decisions at depth $i$ to be of the form $u_i=0$ then $u_i=1$. 
Let $S_c$ be the sum of all the component for which $u_i=1$.
\begin{pte}
The pruning rule
$\forall c_i, w_i + S_c \geq B \wedge u_i = \{0,1\} \implies u_i=0$
applies for the smart design problem.
\end{pte}
Let $ub(u)$ be the upper bound of variable $u$.
The generation function $G$ returns the following learning problem:
\begin{equation}\label{eq:objsmartgen}
\underset{\theta \in \RR^d}{\text{minimize}} \normtwo{X(\theta \cdot ub(u^n))-y} 
\end{equation} 
where $\cdot$ is the Hadamard product.
Since $ub(u^n)$ is a constant vector, problem (\ref{eq:objsmartgen}) can be optimally solved.
Let $\theta'$ be a solution of (\ref{eq:objsmartgen}).
An assignment of $\theta$ of (\ref{eq:objsmart}) is obtained by $\theta = \theta' \cdot ub(u^n)$.
Figure~\ref{fig:ex_123} shows a possible run of the \bl{} framework on this example. 
At root node, the cost is 0.12, which represents a lower bound of the problem.  Then, decision $u_1=0$ is taken and applied.
The next node loss is 0.14. Decision $u_2=0$ is applied.
All the combinations of $u_3$ and $u_4$ are solutions, the current node is a leaf.
Its value is 0.21, the node becomes the current best.
The algorithm backtracks to the previous node and applies decision $u_2=1$.
The best solution is 0.22, which is dominated by 0.21, the node is pruned and the algorithm backtracks to the root.
Decision $u_1=1$ is applied, pruning rules can safely remove 1 from $u_2$ and $u_3$.
All the remaining combination of $u$ are solutions of the problem, and the loss is 0.19. 
This node is the new best.
The algorithm backtracks to the root. 
The optimal value is 0.19.

\paragraph{Example: Prior NMF}
Consider the topic modeling problem using non-negative matrix factorization \cite{NMFreview} from the previous section.
No algorithm, to the best of our knowledge, is able to solve directly such problem. 
Nevertheless, for the vanilla version that does not contains any $ET$ constraints, algorithms exist.
Gradient methods have proven their efficiency for solving such problems.

Consider a \bl{} branching strategy that for each value $j \in s_i$, generate a decision $s_i=j$.
Decision $s_i=j$ implies that $c(w_i,t_j)=0$.
The resulting tree is no longer a binary tree but an n-ary tree.
Consider a search strategy that selects the next decision by selecting the $min_j(\normtwo{w_i \cdot \overline{t_j}})$. 
Let $n$ be the current node, and $D^n=((w_i,t_i),...,(w_j,t_j))$ be the sequence of decisions taken so far.
Let $T^n$ be a matrix having the same dimensions as $W$. 
Each column $T_i^n$ of $T^n$ is defined by $T_i^n = \Sigma_{j \in s_i} t_j$. 
With $\Sigma$ the element-wise logical OR operator.
All the other columns are filled with ones.
The generated problem for node $n$ is:
\begin{equation} \label{eq:objnmfgen}
\underset{W,H}{\text{minimize}} \normtwo{A-(W \cdot T^n) H} 
\end{equation}
This problem too can be solved by gradient method since matrix $T^n$ is not a variable but a constant.

\begin{algorithm}[tb]
\caption{\bl{} algorithm \label{a:branchAndLearn}}
\label{alg:algorithm}
\textbf{Input}: $T_d$\\
\textbf{Output}: $BestNode$
\begin{algorithmic}[1] 
\STATE root $\leftarrow$ Node()\\
\STATE $L \leftarrow$ Set(root)\\
\WHILE{$L \neq \emptyset$ and $shouldStop() = False$}
\STATE 	$n \leftarrow selectAndRemove(L)$\\
\STATE 	$prune(n)$\\
\STATE 	$generateModel(n)$\\
\STATE 	$train(n)$\\
\STATE 	$L.append(generateDecision(n))$\\
\STATE 	$keepBest(n)$
\ENDWHILE
\end{algorithmic}
\end{algorithm}

\noindent\textbf{Link with combinatorial solvers}
The \bl{} framework can, and should, easily be implemented in most existing combinatorial solvers such as constraint programming (CP).
Moreover, using existing CP solvers directly provides us strong combinatorial optimization power, allowing us to focus on the learning part. 
The learning part can be encapsulated as a constraint directly. 
Indeed, the branch and bound and its three components are already the core of such solvers, only the generation and learning parts are missing.
This constraint should be run once the propagation is finished, which implies generating the learning problem as a function of the state of the CSP, training the model, and extracting the loss.
If the stopping conditions related to learning are reached, then a fail is triggered, otherwise, the CP solver continues its work.

\section{Related work}
In many cases, the \bl{} principles are going to be relatively similar to existing \bb{} problems, such as the search strategy which will still select a given sub-problem from a set of possible ones. 
Thanks to that, we will be able to re-use the huge work on black-box optimization, especially for the modeling part (constraint language, etc.).
The major difference leans in the pruning rules, the use of the objective function from the model, and the interactions between the \bb{} variables and the learning process (i.e. the learning model generation).
In our problem, the objective function is often going to be the same as the machine learning model that we try to optimize.
The use of \bb{} algorithms has already been used in machine learning and feature selection, thus searching onto the L0 constraint \cite{narendra77,chapelle07}. 
Our work generalize such works and allows not only to search onto the domain of the variables, but also onto the constraint set, because we can have higher level variables (i.e. searching over the table points). 
Many works are now using machine learning to improve the branch and bound, and optimization in general \cite{selsam18,haim09}, while these are promising works, this paper aims to do the opposite, to use \bb{} to improve the consistency of machine learning models.
Recently, the use of combinatorial optimization solver for machine learning has been done in the context of target moving \cite{detassis21}. 
Their work iterates between a pure machine learning phase, and a pure combinatorial phase to change the target of the optimization. 
This is different from our work, we propose use the combinatorial part to restrict the hypothesis set of the machine learning part.
Learning models are often used inside of CP solvers, either to approximate/learn constraints, or as objective function etc.
In all of these works, the machine learning models are pre-trained, and the parameters are fixed \cite{lombardi2017empirical}.
Table constraints and cost is a known topic in constraint programming, where soft implementation are used to model over-constrained problem \cite{khong2017efficient}. 
Our settings are different as we generalize such work and embed different distance function.
Finally, the proposed work differ from methods iterating between applying different optimization direction to the current point \cite{condat2013primal,marra2020local}. 
The proposed method generates each time a new sub-problem which is optimized, instead of shifting the current solution.

\section{Experiments}
The experimental section is split in two parts. 
The first part is about smart design.
The second part is about the NMF with prior.
This section shows how the \bl{} framework leads to high quality solutions for these constrained learning problems. 
Both models have been implemented in the same \bl{} framework implementation.

\subsection{Smart Design instances}
We generated several hundreds instances with the number of feature sizes $n$ being in the list (10, 20, 40, 70, 100, 130, 150, 180, 200, 225, 250, 300, 350), and the number of samples $T$ being in the list (100, 400, 700, 1000, 1500, 3000, 7000, 10000), and the percentage of total weight $C_{\%}$, used to define the max cost $C$ value, in the list (0.90, 0.80, 0.60, 0.30).
All of the instances contains an additional Gaussian noise.
We set a time out of 10 minutes.
All of the source code can be found on Github\footnote{Hidden link}. 
As competitors, we defined two methods. 
They are based on the regression optimization problem of equation (\ref{eq:objsmart}).
The first one is to apply classic linear regression first, and
then to remove the smallest coefficients until the constraint is respected. 
We then fit the model again, with the selected subset of features.
We call this algorithm $L2_{BR}$ for Basic Repair. 
The second one use the information of the weights of the input by computing the ratio coefficient over weight. 
We call this algorithm $L2_{OR}$ for orthogonal repair by analogy with orthogonal basis pursuit.
In order to evaluate the algorithms, we split our data-set into a training set of size 80\% and a test set size of 20\% used for evaluation and applied 5 folds.
All the accuracy showed here are with respect to the test set, otherwise it will be explicitly said.

\begin{figure}[h]
\centering
\includegraphics[width=\figsize]{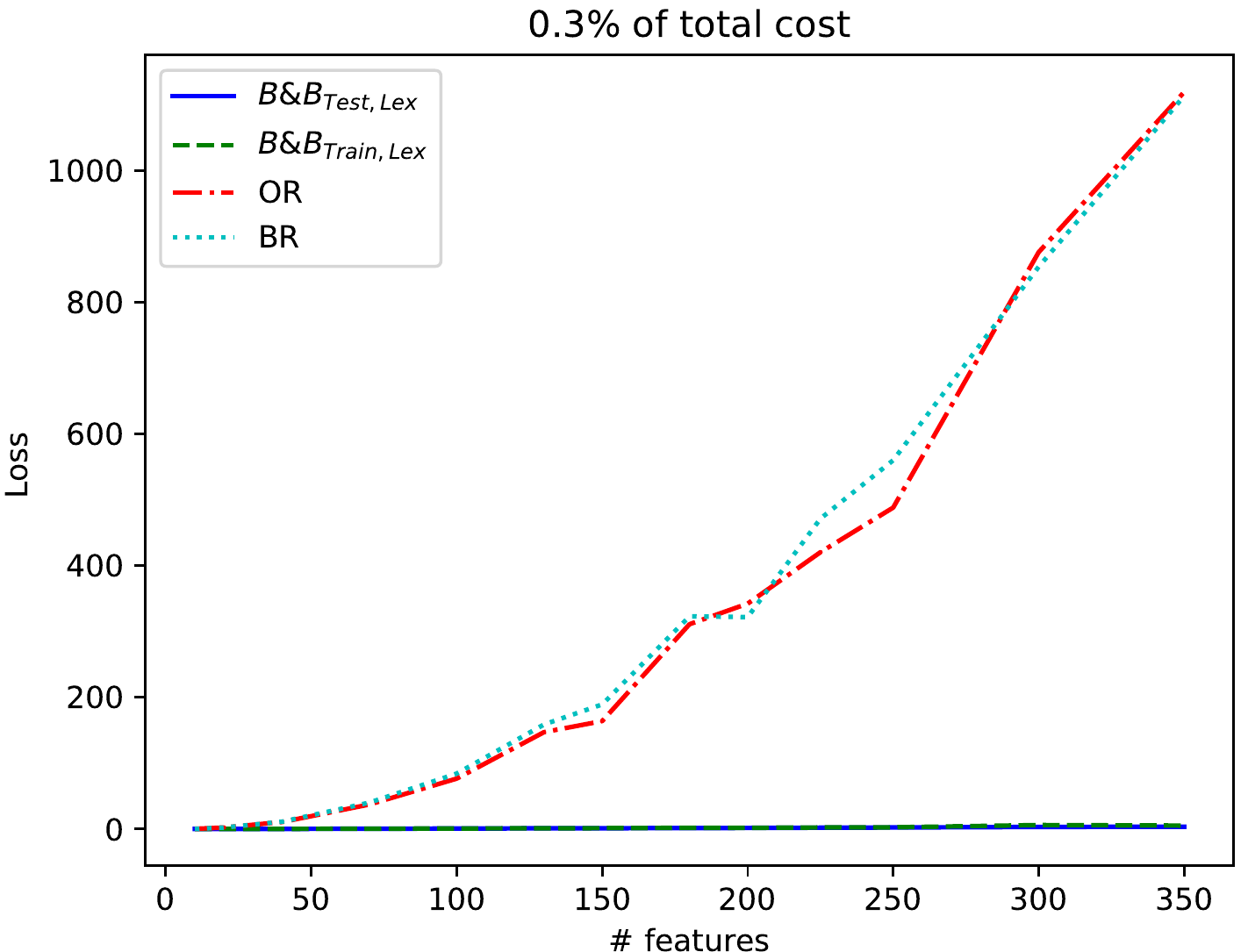}
\includegraphics[width=\figsize]{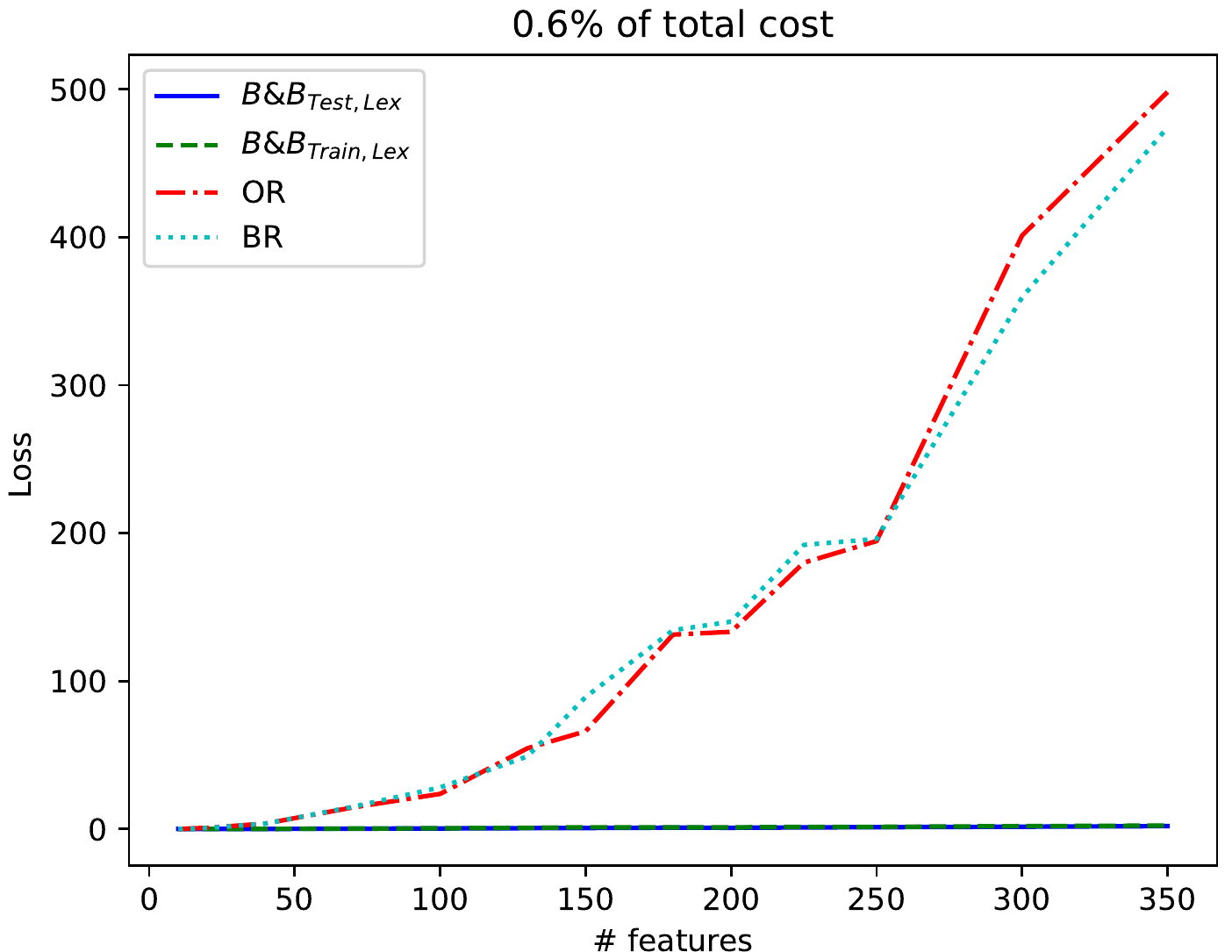}
\includegraphics[width=\figsize]{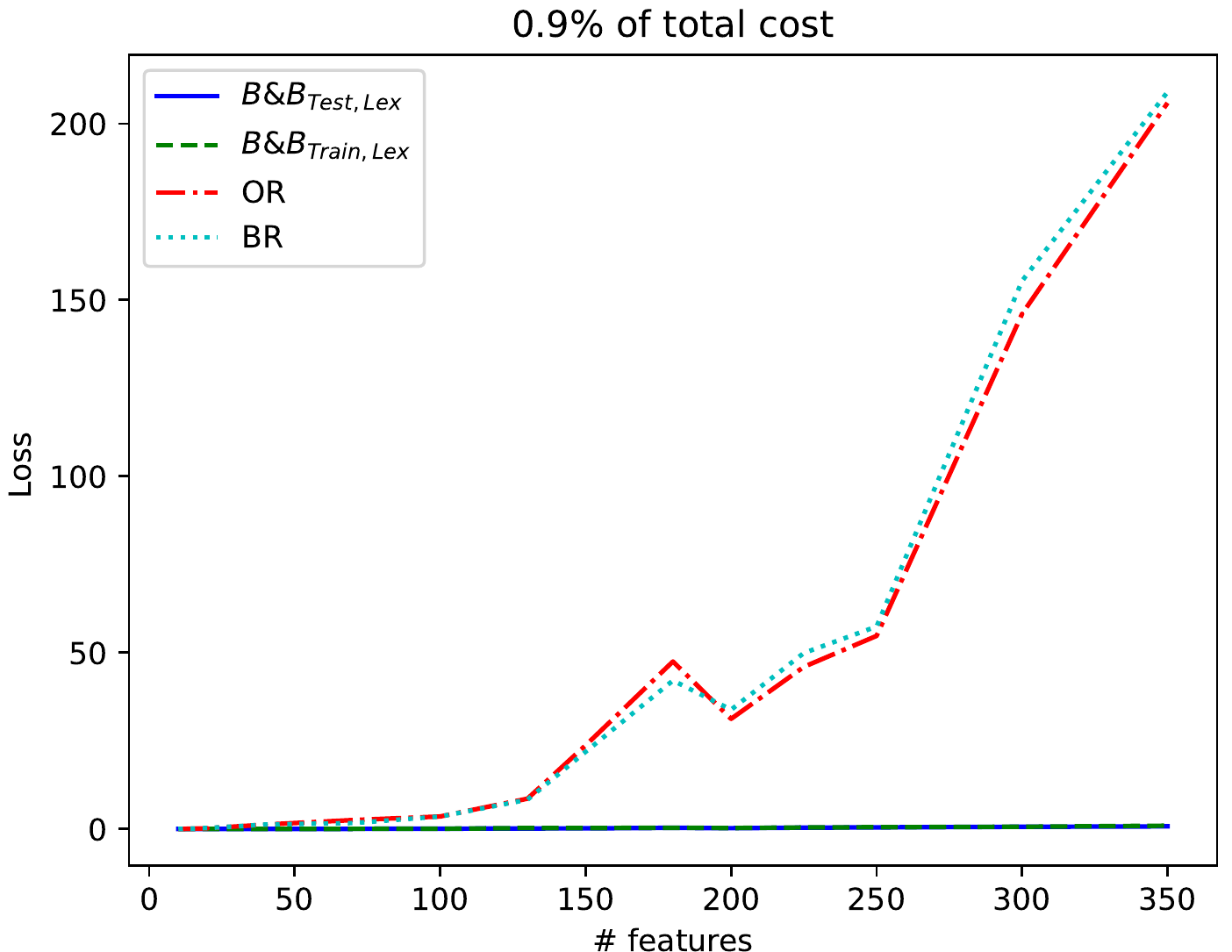}
\includegraphics[width=\figsize]{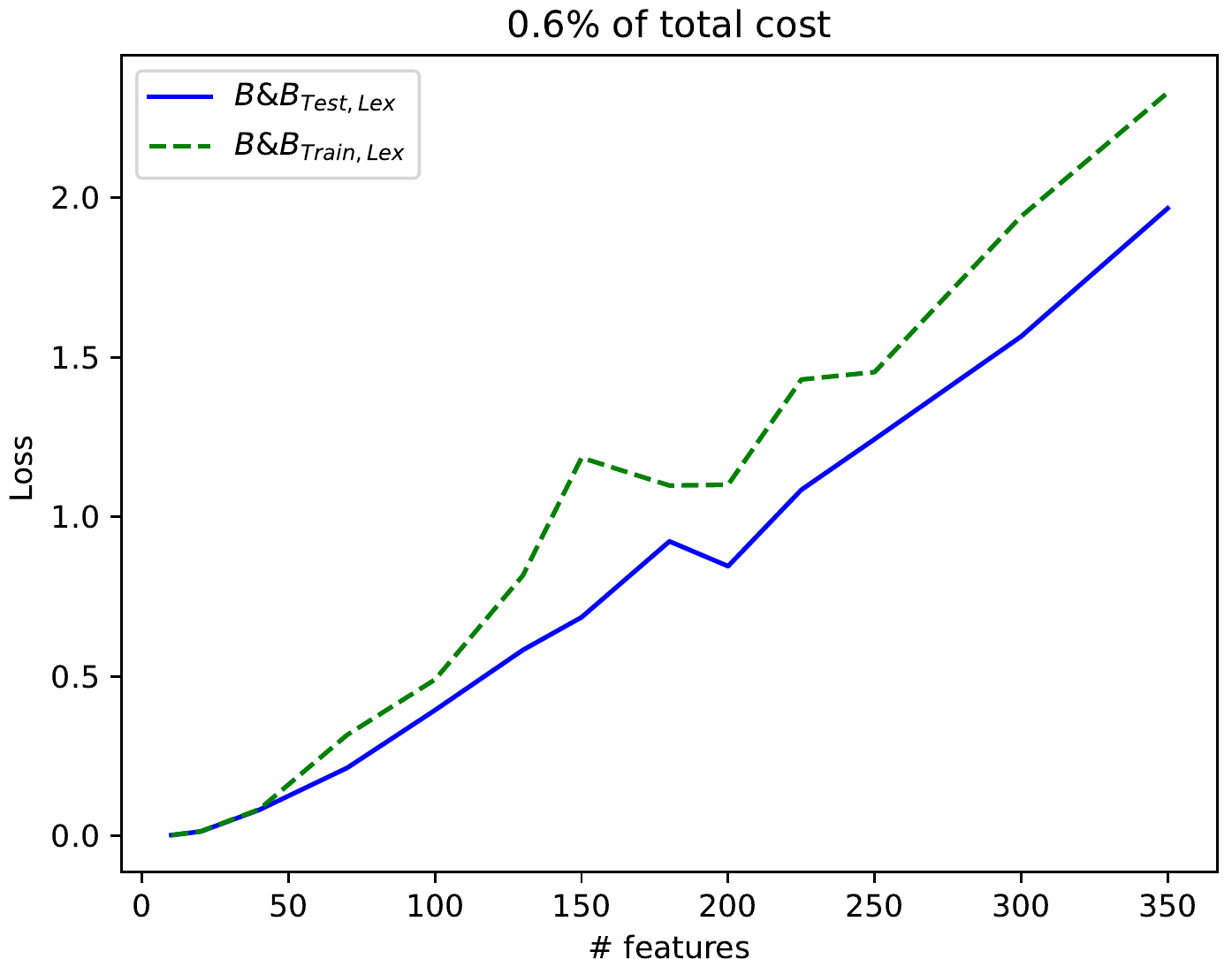}
\caption{Mean loss value for a given percentage of the maximum loss.} 
    \label{fig:percentImpact}
\end{figure}

\begin{figure}[h]
\centering
\includegraphics[width=\figsize]{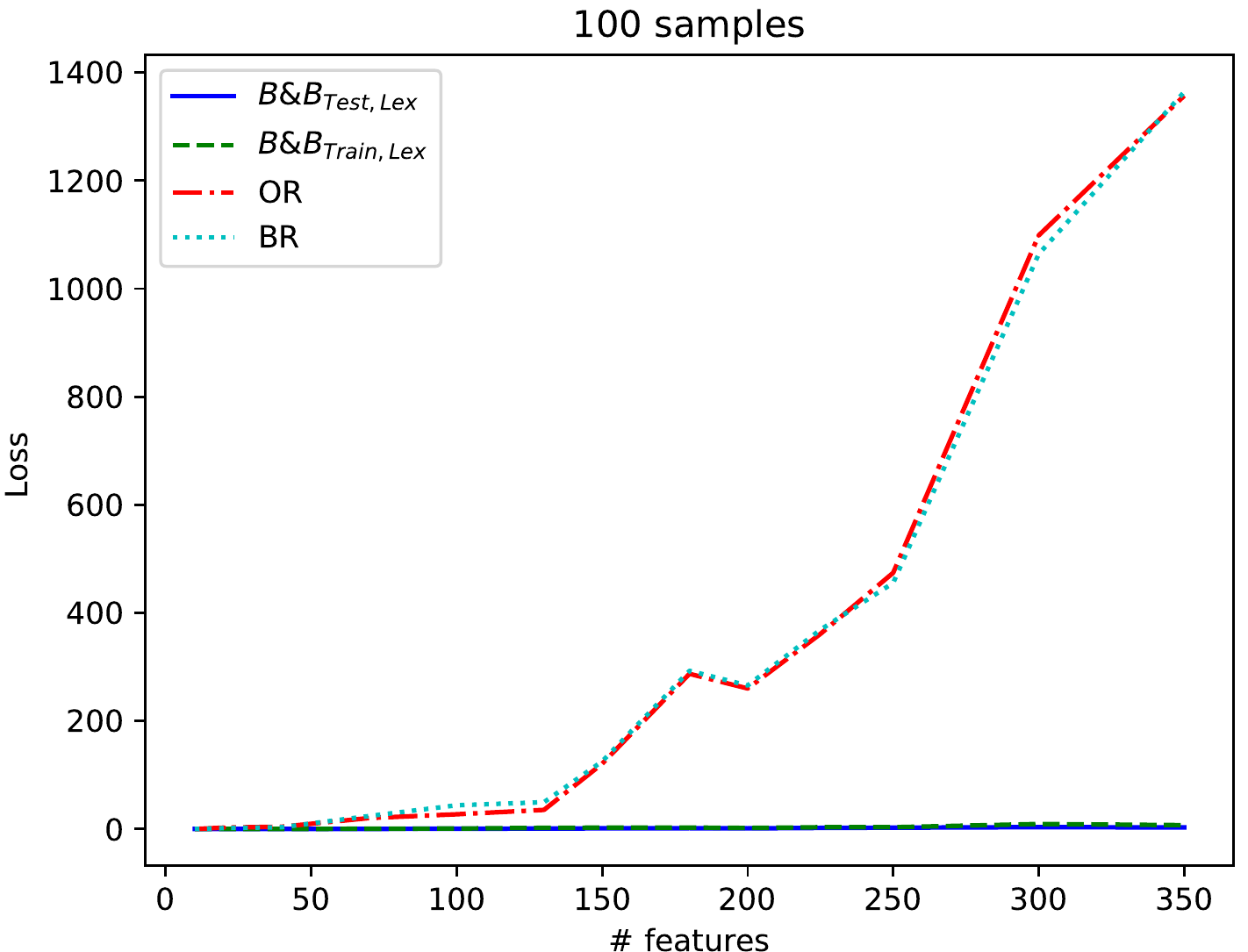}
\includegraphics[width=\figsize]{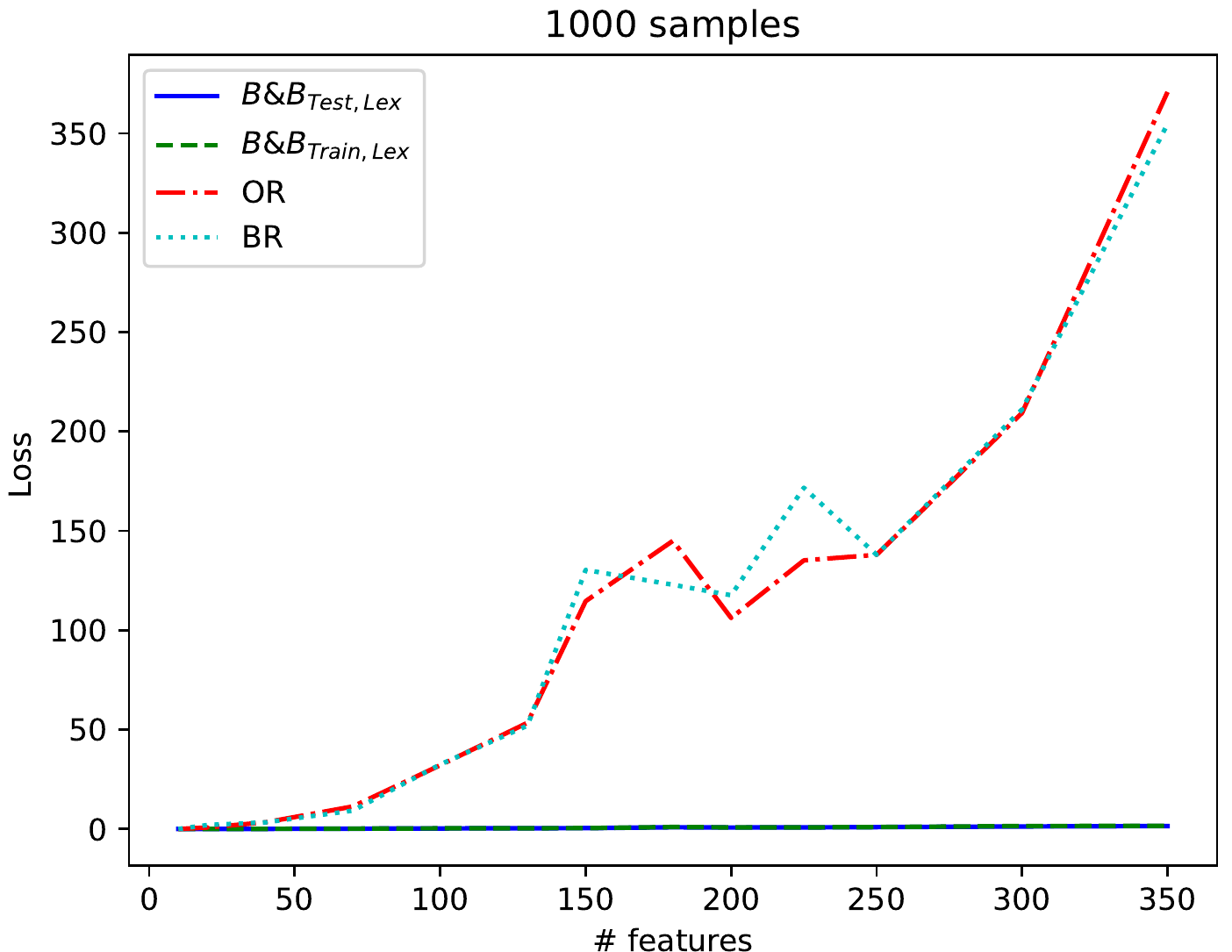}
\includegraphics[width=\figsize]{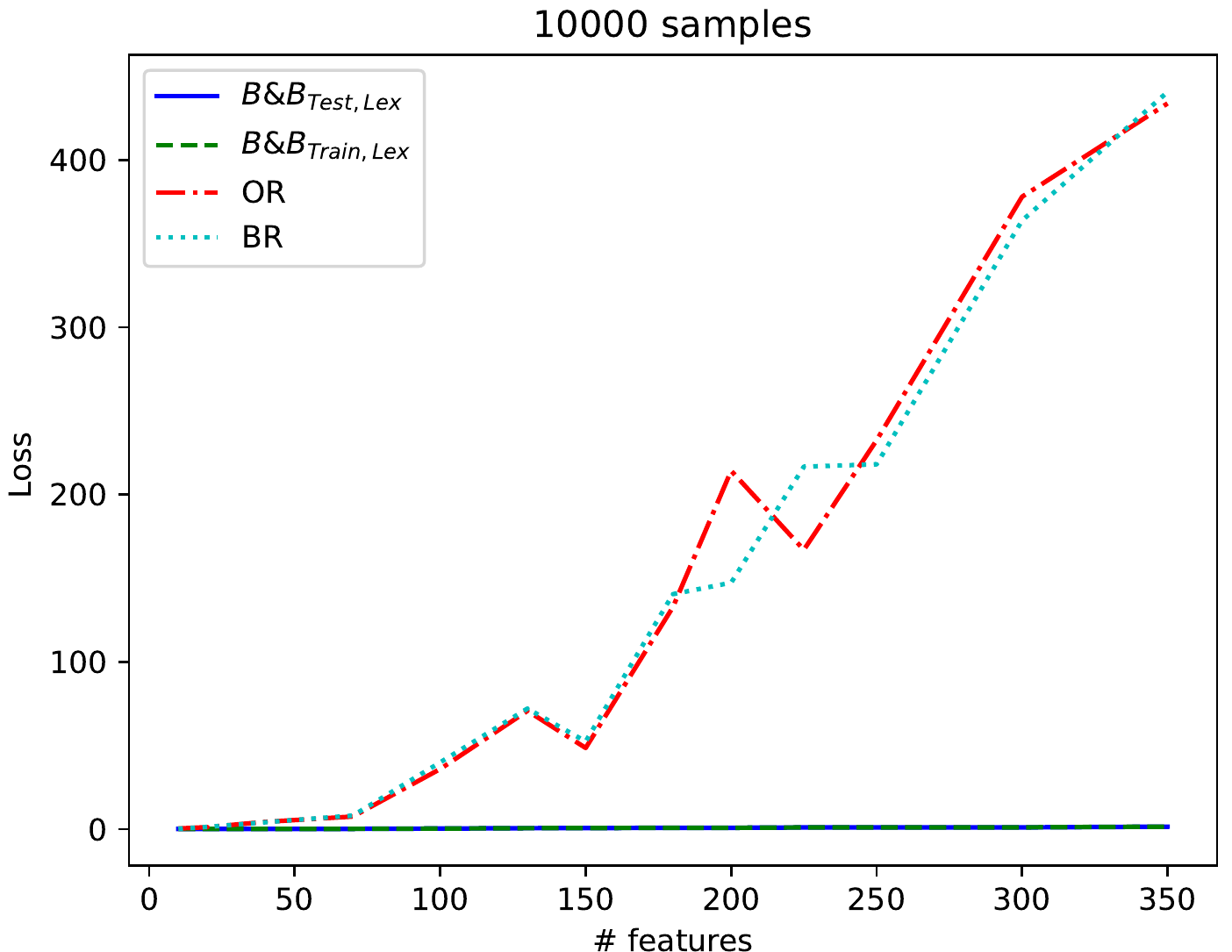}
\includegraphics[width=\figsize]{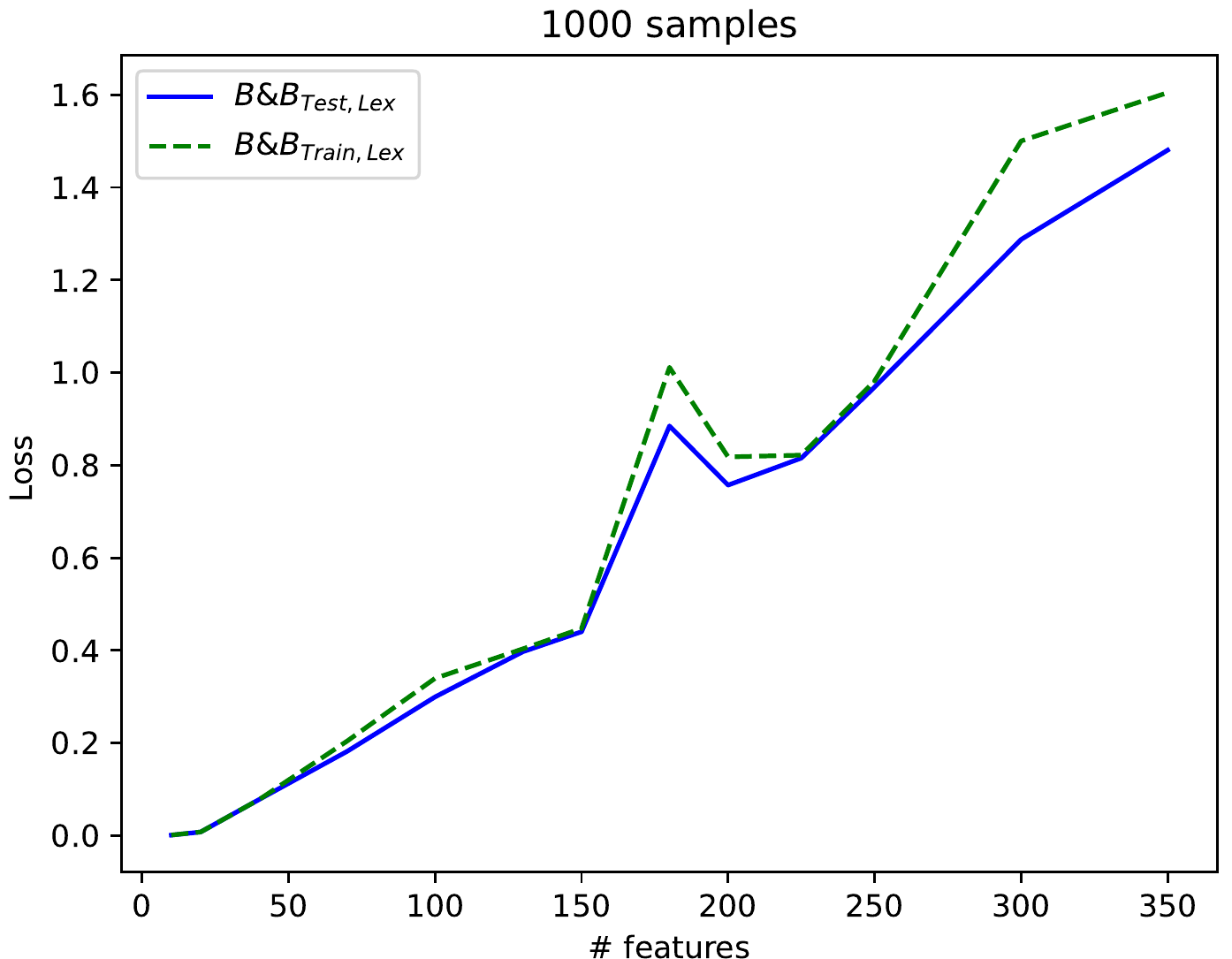}
\caption{Mean loss value for a given number of samples.} 
    \label{fig:sampleSizeImpact}
\end{figure}

\textit{Loss.} Figure~\ref{fig:percentImpact} shows the test loss of all the methods for different values of the $B$ (maximum cost) variable.
As we can see, the \bl{} framework results are strictly stronger than the $L2_{BR}$ and $L2_{OR}$ methods.
Such a result is not surprising as the \bl{} will search for the best subset of component that maximize the loss.
While the $L2_{BR}$ and $L2_{OR}$ results are the component that are the most used or whose ratio is maximum in the optimal solution of the unconstrained fitting problem. 
Figure~\ref{fig:sampleSizeImpact} shows the test loss of all the methods for different with respect to different number of samples. The less we have samples, the harder it is to fit our model.
From this figures, we can see that either with respect to the constraint bound or the number of samples, the \bl{} framework seems is able to extract better solutions than the reconstruction methods.

\begin{figure}[h]
\centering
\includegraphics[width=\figsize]{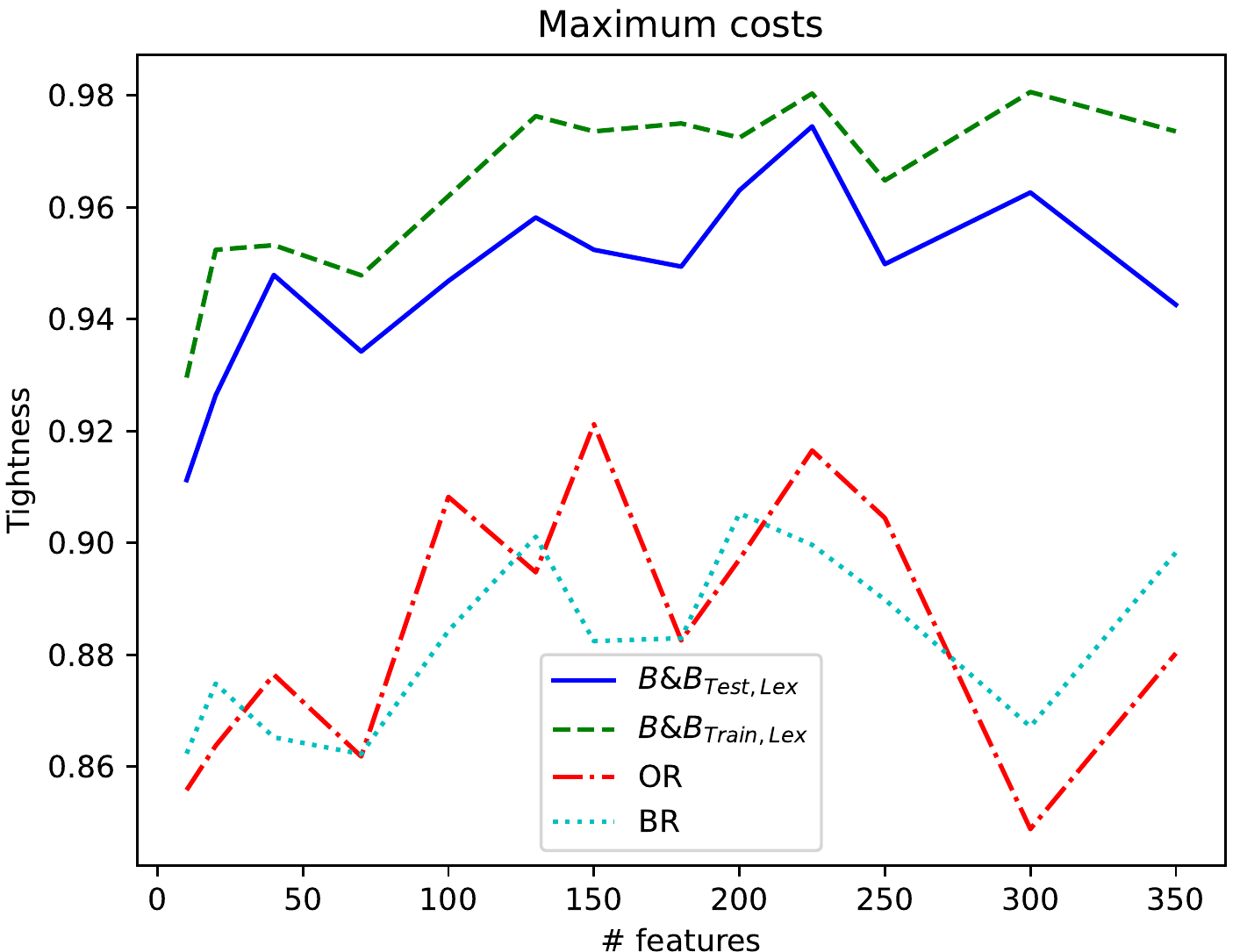}
\includegraphics[width=\figsize]{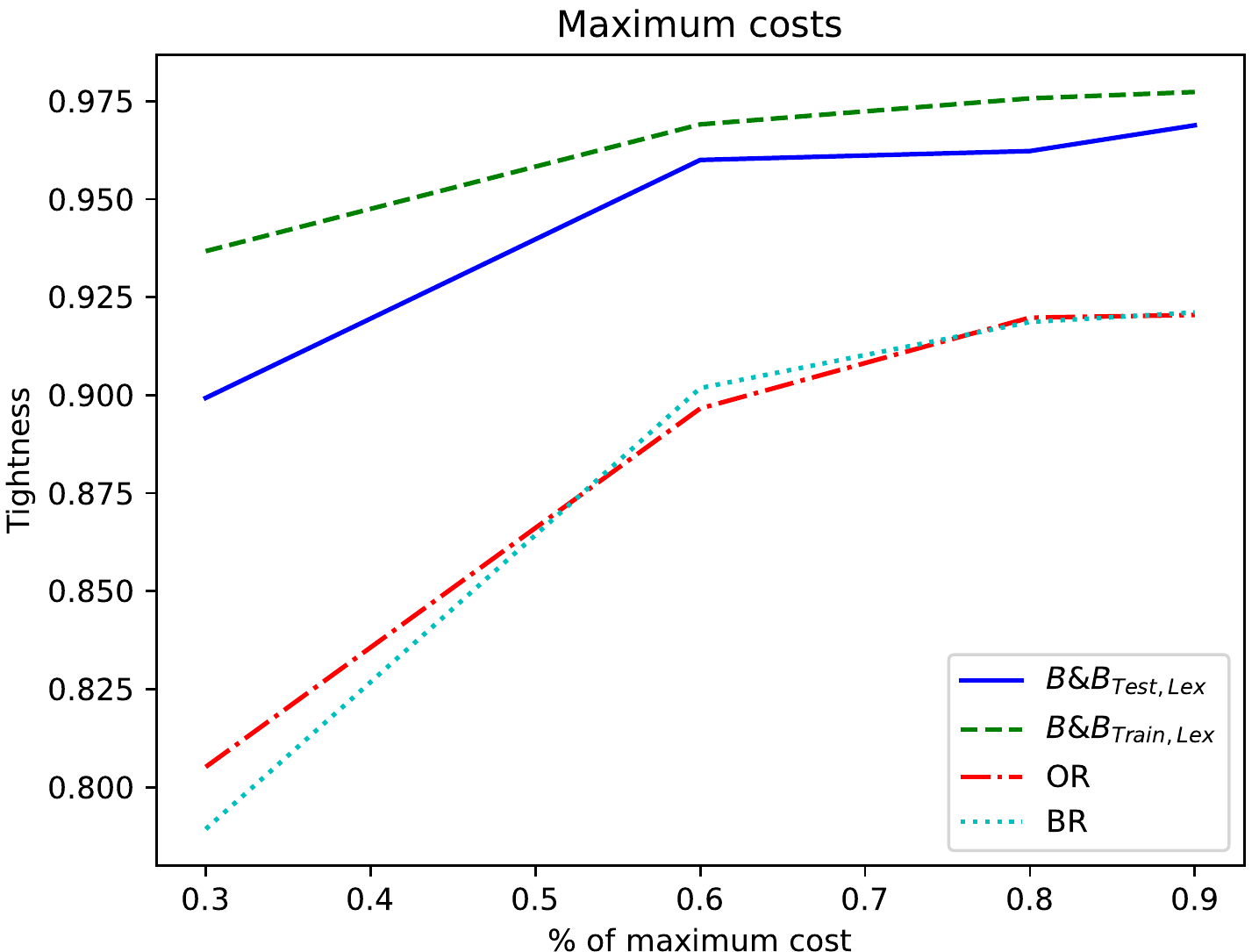}
\caption{Tightness of the constraint satisfaction for the maximum cost constraint.} 
    \label{fig:MaxCostPlot}
\end{figure}

\textit{Constraint.}
Figure~\ref{fig:MaxCostPlot} shows the percentage of the tightness of the satisfaction of the maximum cost constraint. 
This is given by the ratio $\frac{\sum_{i}^{C} u_iw_i}{B}$ (i.e. actual cost divided by maximum cost).
As we can see, the \bl{} framework seems to tighten the constraint as much as possible, and in most experiments, there is a gap of around 10 \% between the \bl{} algorithm and the reconstruction methods.
Such a result is interesting as it is intuitively important that being less impacted by the constraints should give more freedom for the learning part. 
Even if in many cases, the introduced constraints can be incorporated for helping the learning process.

\begin{figure}[h]
\centering
\includegraphics[width=\figsize]{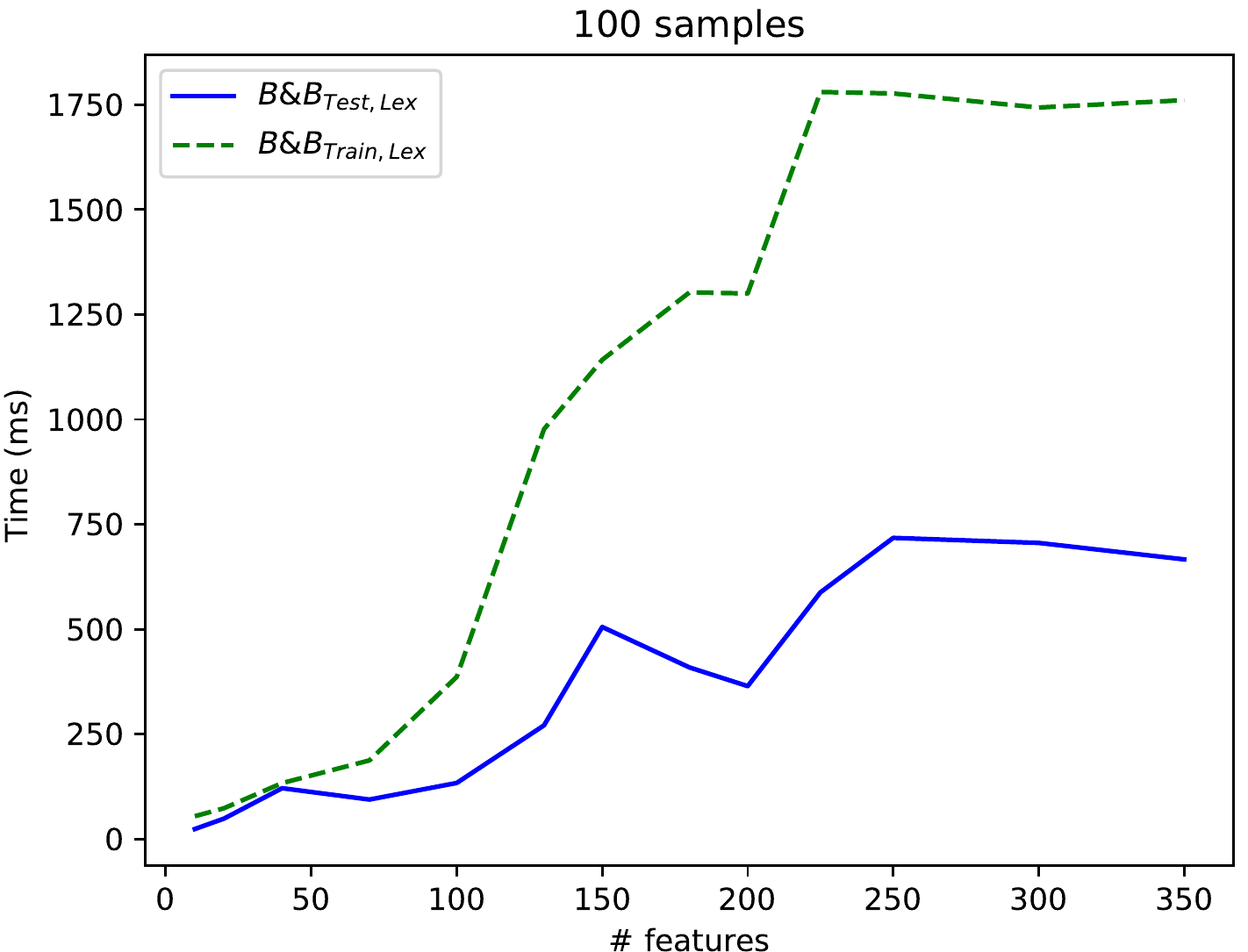}
\includegraphics[width=\figsize]{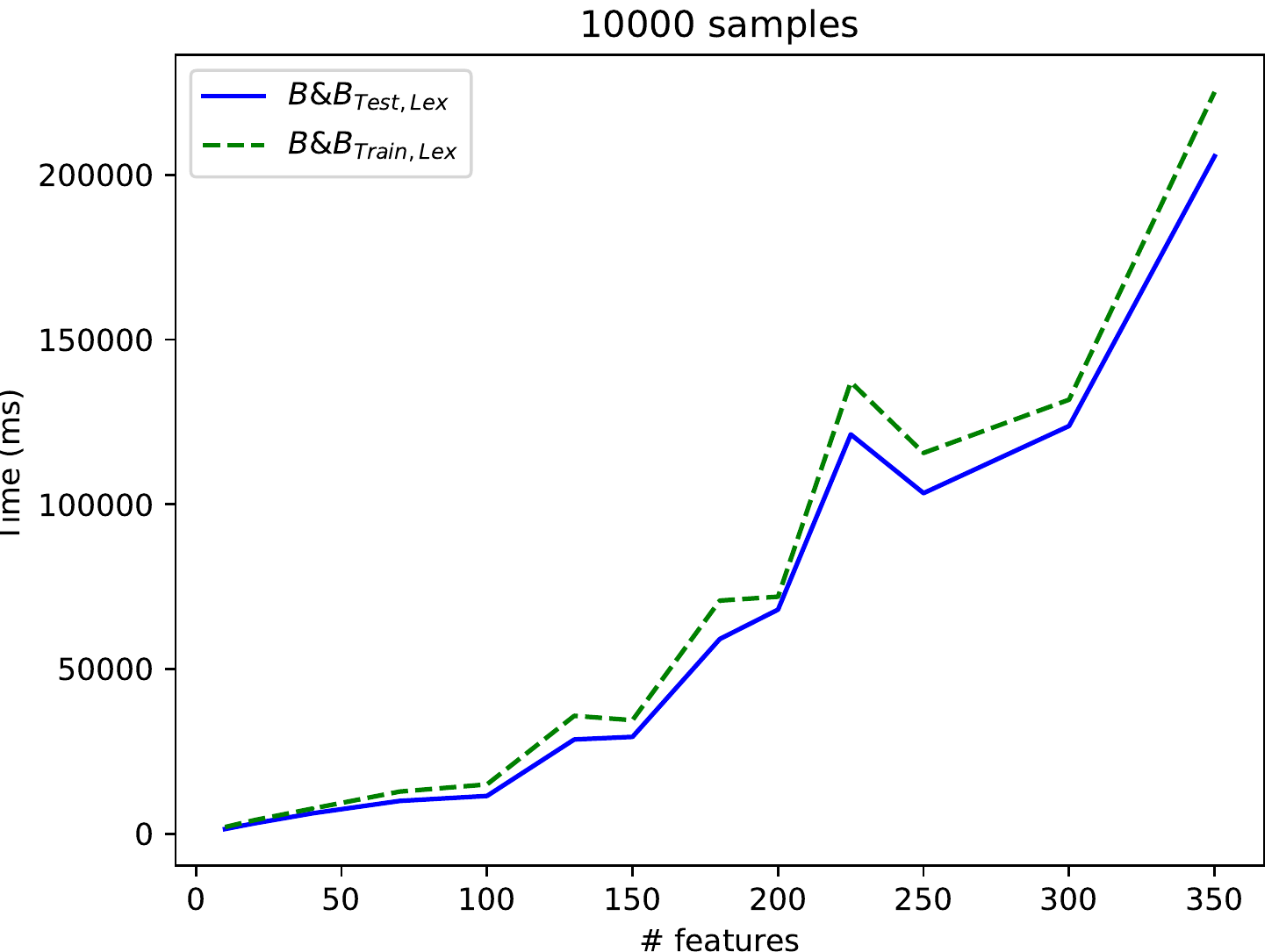}
\includegraphics[width=\figsize]{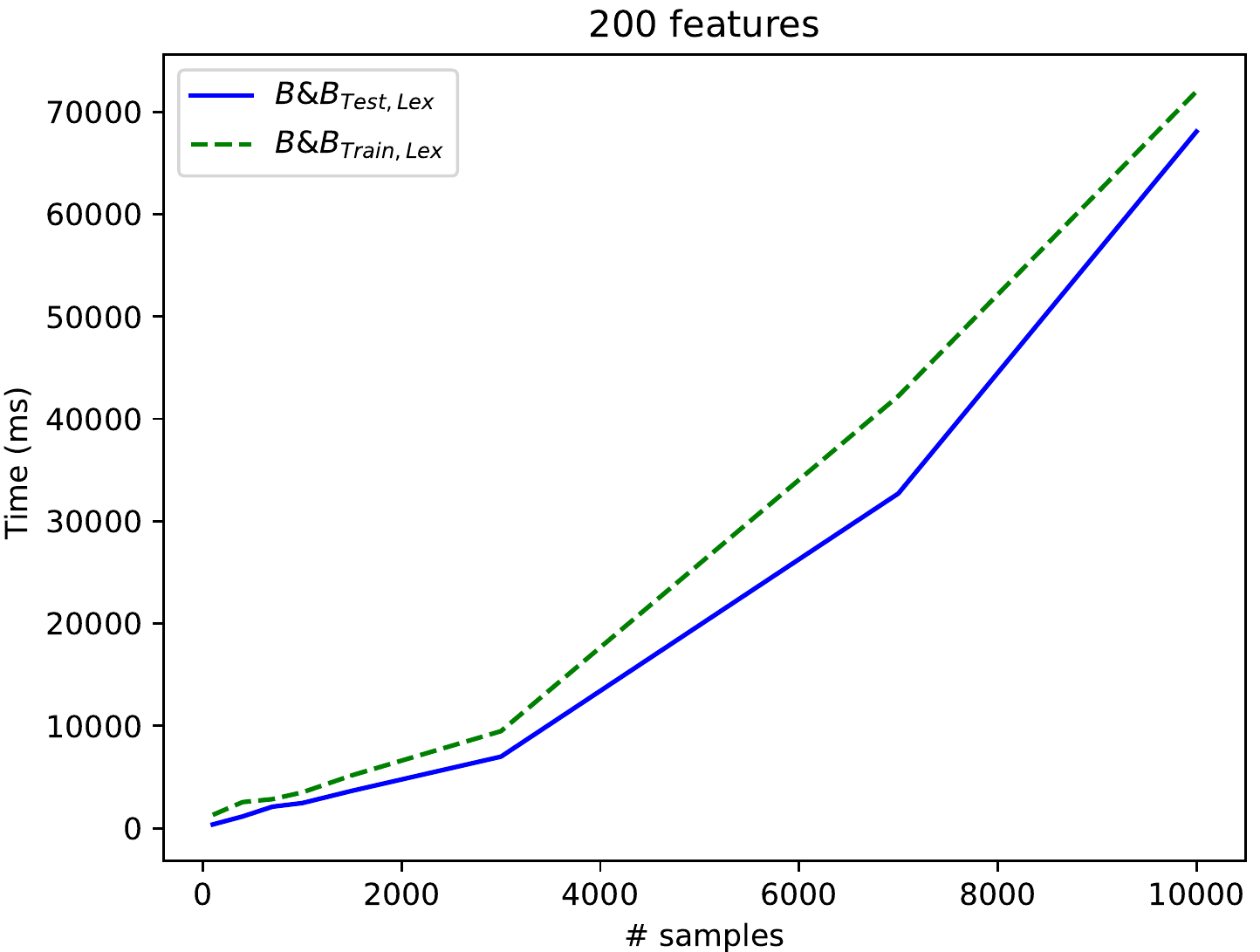}
\includegraphics[width=\figsize]{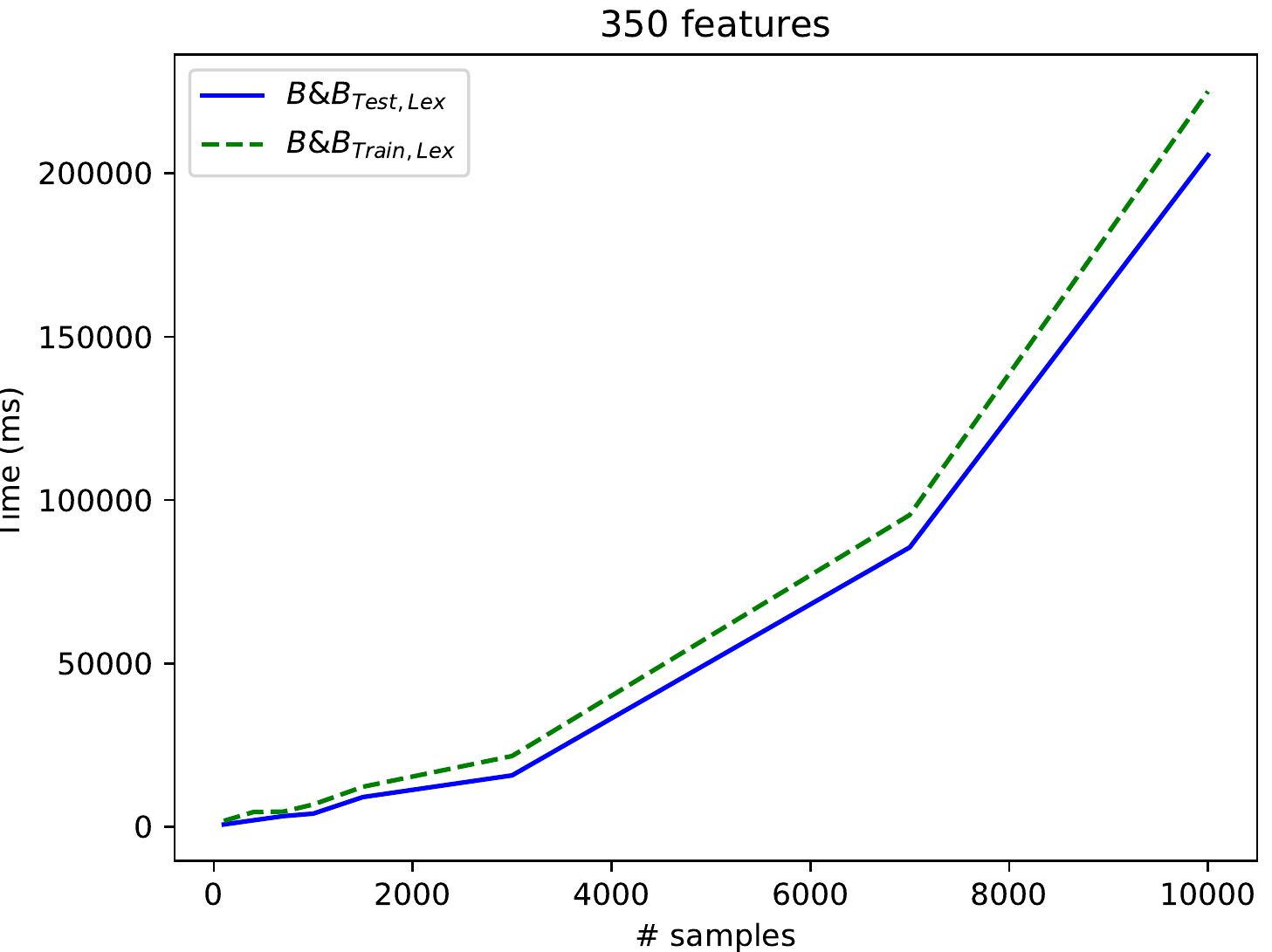}
\caption{Mean time for a given number of samples, or given a number of features.} 
    \label{fig:TimeSample}
\end{figure}

\begin{figure}[h]
\centering
\includegraphics[width=\figsize]{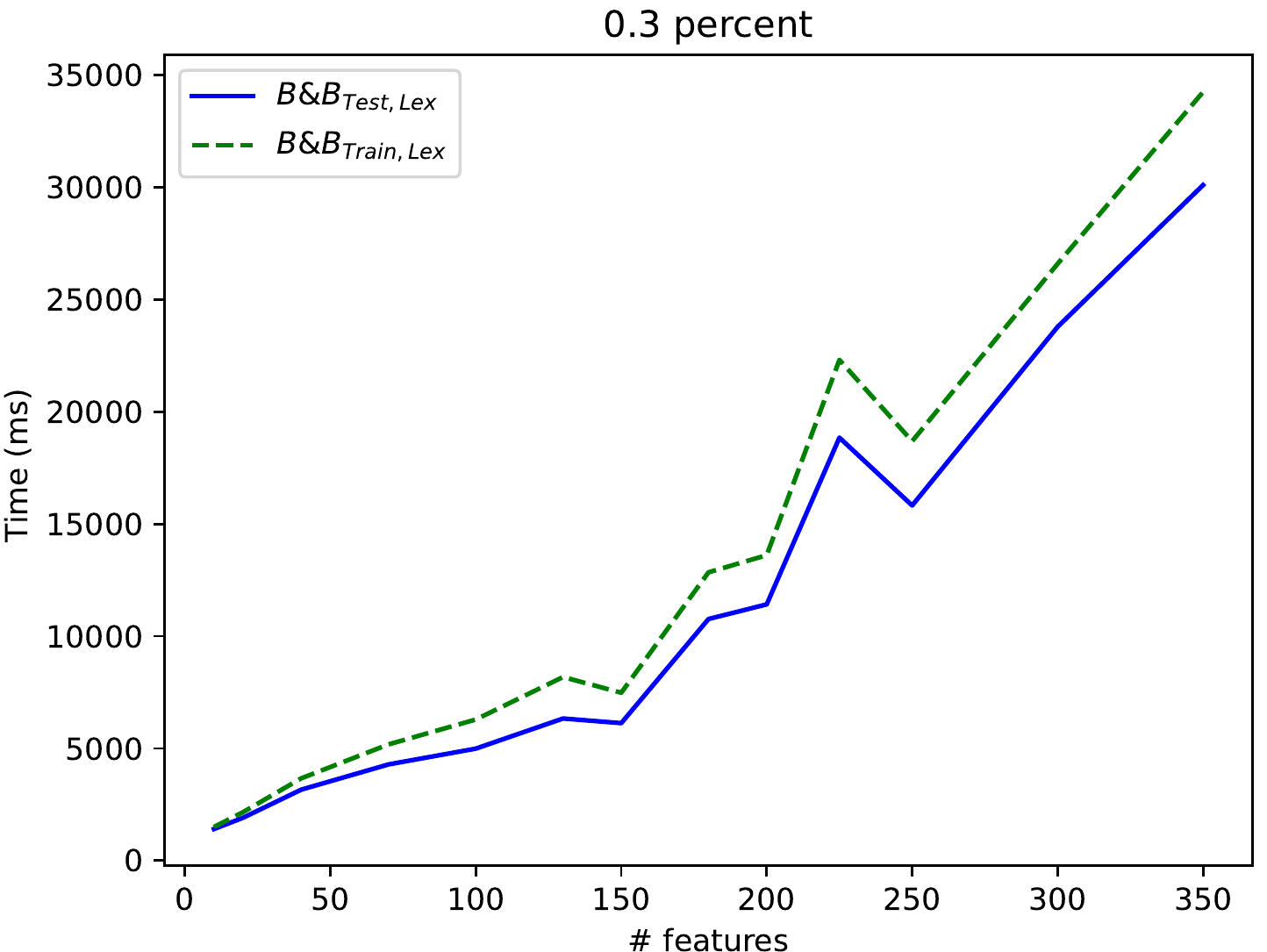}
\includegraphics[width=\figsize]{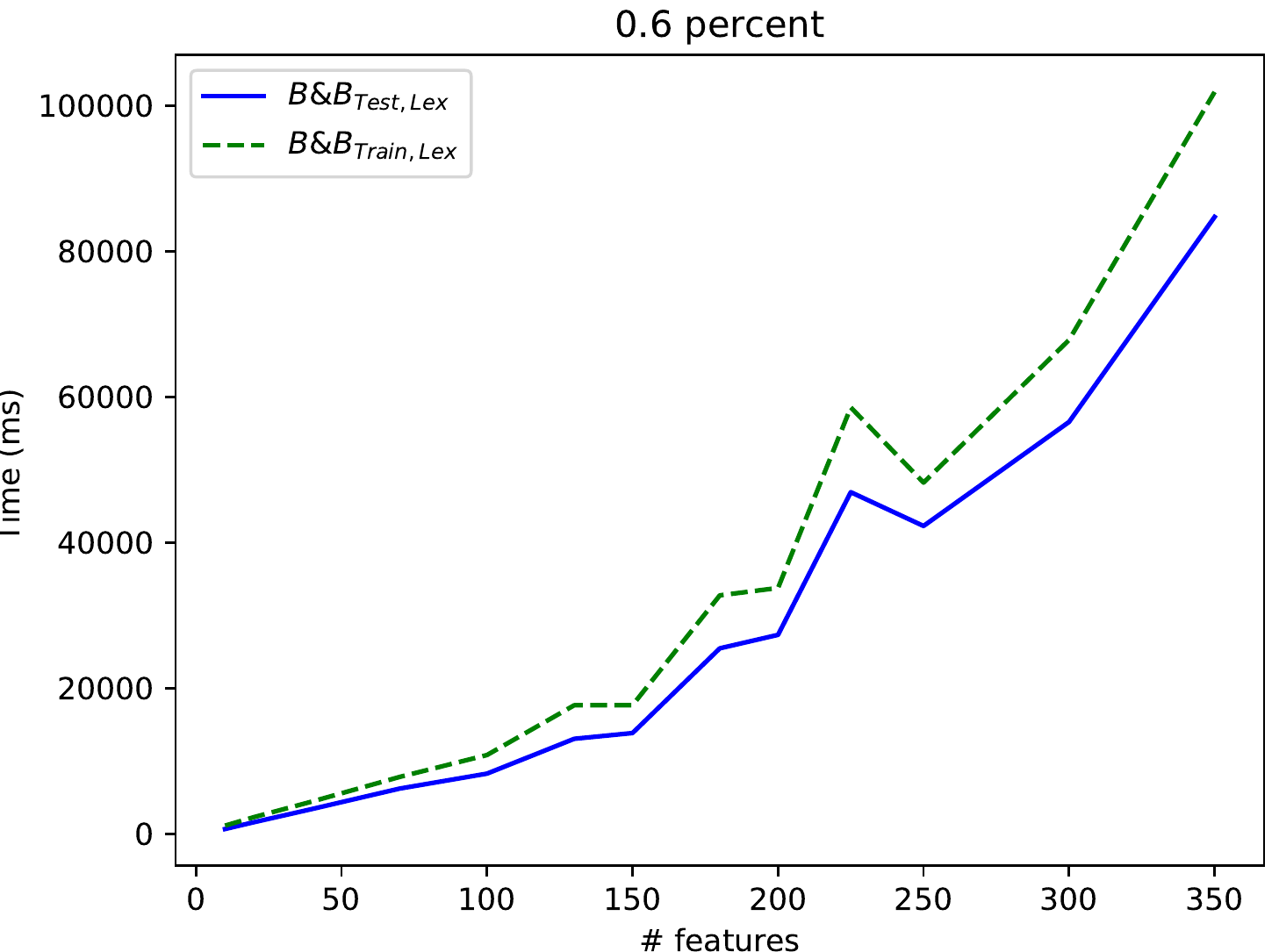}
\includegraphics[width=\figsize]{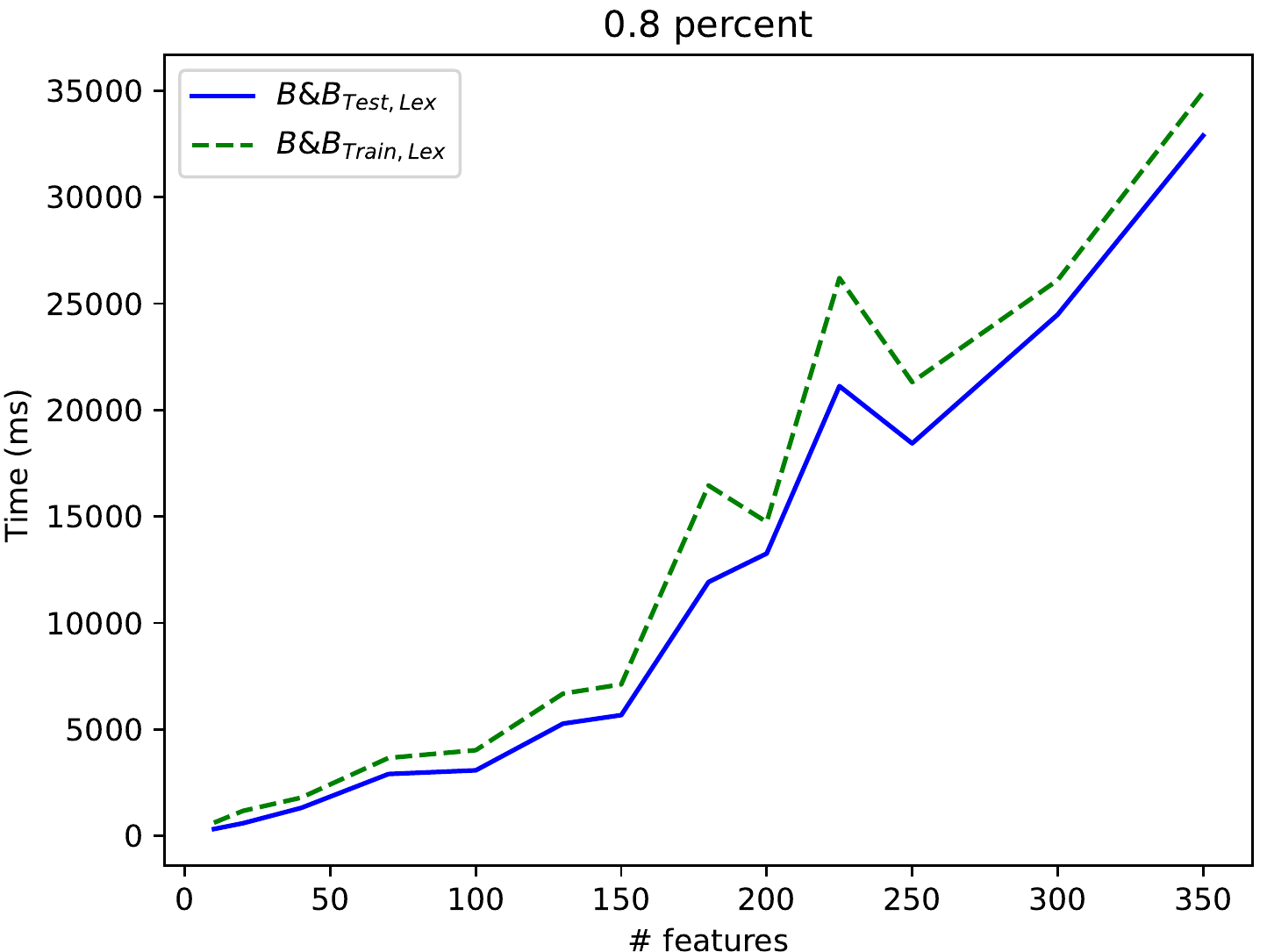}
\includegraphics[width=\figsize]{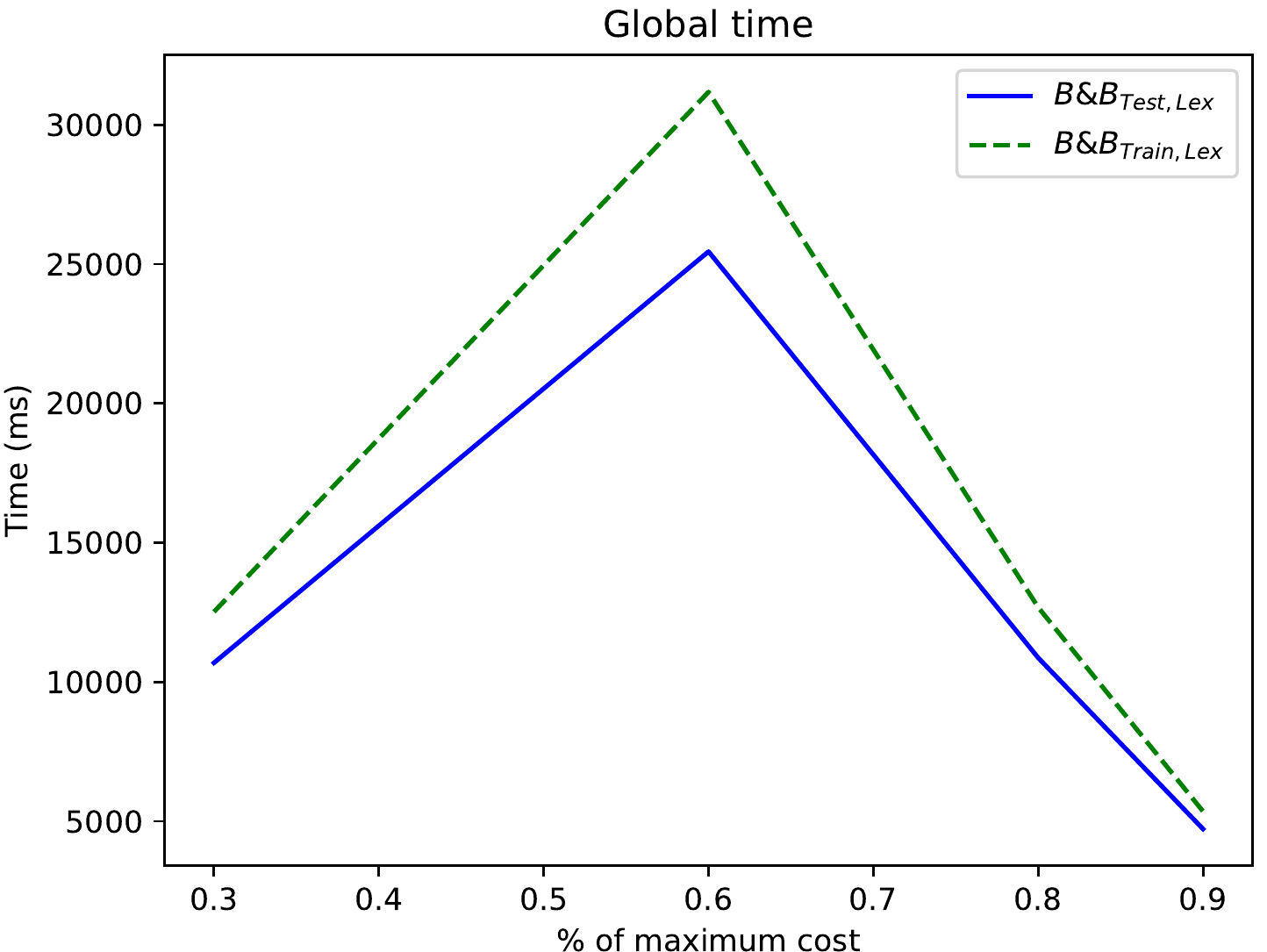}
\caption{Mean time with respect to the maximum cost constraint.} 
    \label{fig:TimePercent}
\end{figure}

\textit{Time.}
The last important factor to analyze is the time. 
For the reconstruction methods, the time will grow with the number of features and number of samples, but compared to the \bl{} time, the $L2_{BR}$ and $L2_{OR}$ running time is insignificant most of the time.
Figure~\ref{fig:TimeSample} shows the impact of the number of samples and number of features on the running time of the \bl{} framework.
As we can see, the time consumption growth with the number of features and the number of samples drastically. 
In combinatorial problems, it is not usual to have exponential running times like, as they are used to solve NP-Hard problems. %
In most large scale instances, solving the complete tree search will be computationally infeasible.
As for classic \bb{} problems, smart search strategies and decomposition methods should be defined to scale up.
Figure~\ref{fig:TimePercent} shows the impact of the constraint on the running time.
As we can see, the longest running times are neither for the smallest value of the constraint, or for the largest, but for a value in the middle. Here 0.6 percent of the total costs.
This result could be explained by the following: when the constraint is too strong, the search space is strongly cut by the constraint. When the constraint is loose, good solutions can be found easily and allows to cut the search space too.
Such results in inherent in combinatorial optimization solvers, where adding constraints to solve a problem faster is often done \cite{gomes04}.

\subsection{Prior database topic modeling}
The filtering of the \textit{allDifferent} constraint is a binary one.
We generated random instances matching these settings (i.e. having a degree of novelty of $t$ with respect to a database of knowledge).
The number of words $n$ is in the list (20, 30, 50, 75, 100, 150).
The number of true topics $T_t$ per instances is in the list (4, 5, 6, 7, 8).
The number of additional false topics $F_t$ to be added in the prior database is in the list  (2, 3, 5, 10)
The number of documents $m$ per instances is int he list (50, 100, 150, 200, 250, 300),
The database of prior $T$ is created by removing $t$ topics from the topics involved in an instance and concatenating with the false topics.
In addition, we set the sparsity to be 0.8 for both matrices $H$ and $W$. The minimal number of topics per document is 2.
All of the instances contains an additional Gaussian noise.

The purpose of such an experiment is to check if a hard implementation of a prior knowledge using combinatorial search can lead the learning process. 
The NMF solver uses multiplicative gradient updates \cite{NMFMultStat}, and all of the source code can be found on Github\footnote{Hidden link}.
Such an implementation has the advantage of being unable to change a value initially set to zero by the product with matrix $T^n$.
The search strategy chooses the next point on the prior database by selecting the closest point of the current column of $W$ with respect to its L2 norm.

\begin{figure}[h!]
\centering
\includegraphics[width=7cm]{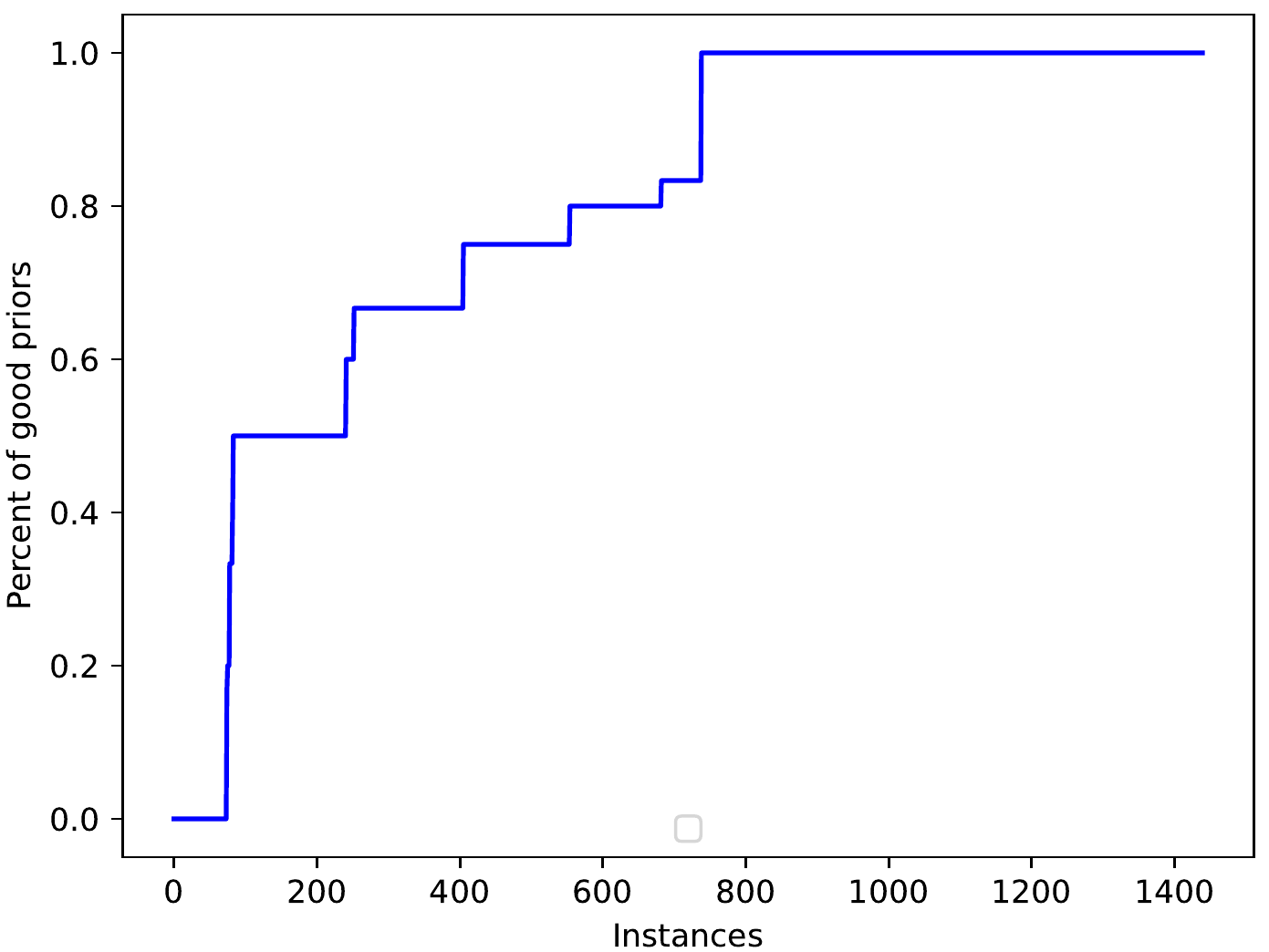}
\caption{Percentage of correct selection of priors in the best solution found.} 
    \label{fig:nmfPriorUse}
\end{figure}

Figure~\ref{fig:nmfPriorUse} shows the percentage of topics correctly extracted from the database of prior knowledge. 
As we can see, most of the time, the true topics are extracted from the prior knowledge DB.
A deeper investigation showed that in several instances where some false topics where used, they allowed to model the noise.

\begin{figure}[h]
\centering
\includegraphics[width=\figsize]{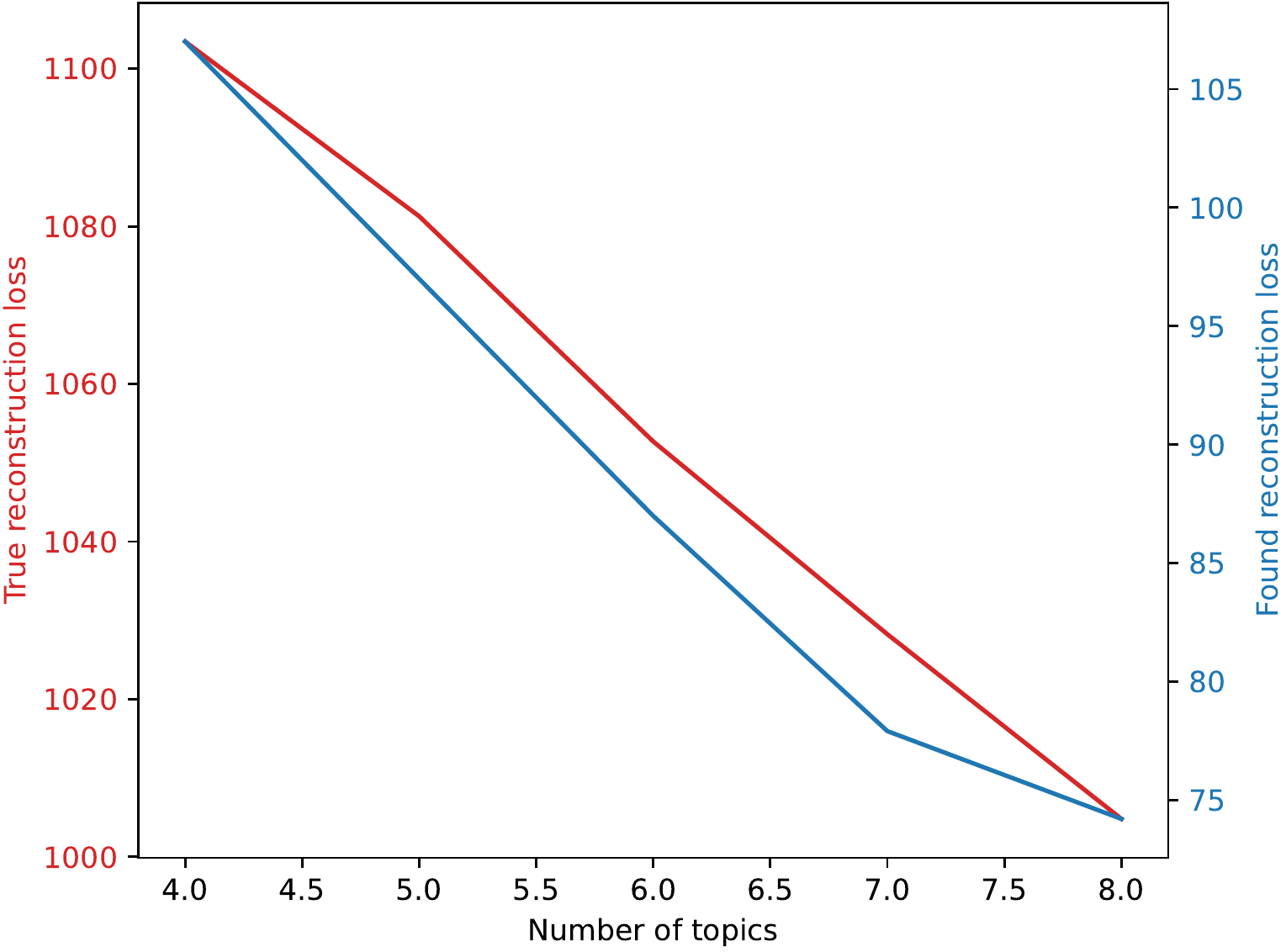}
\includegraphics[width=\figsize]{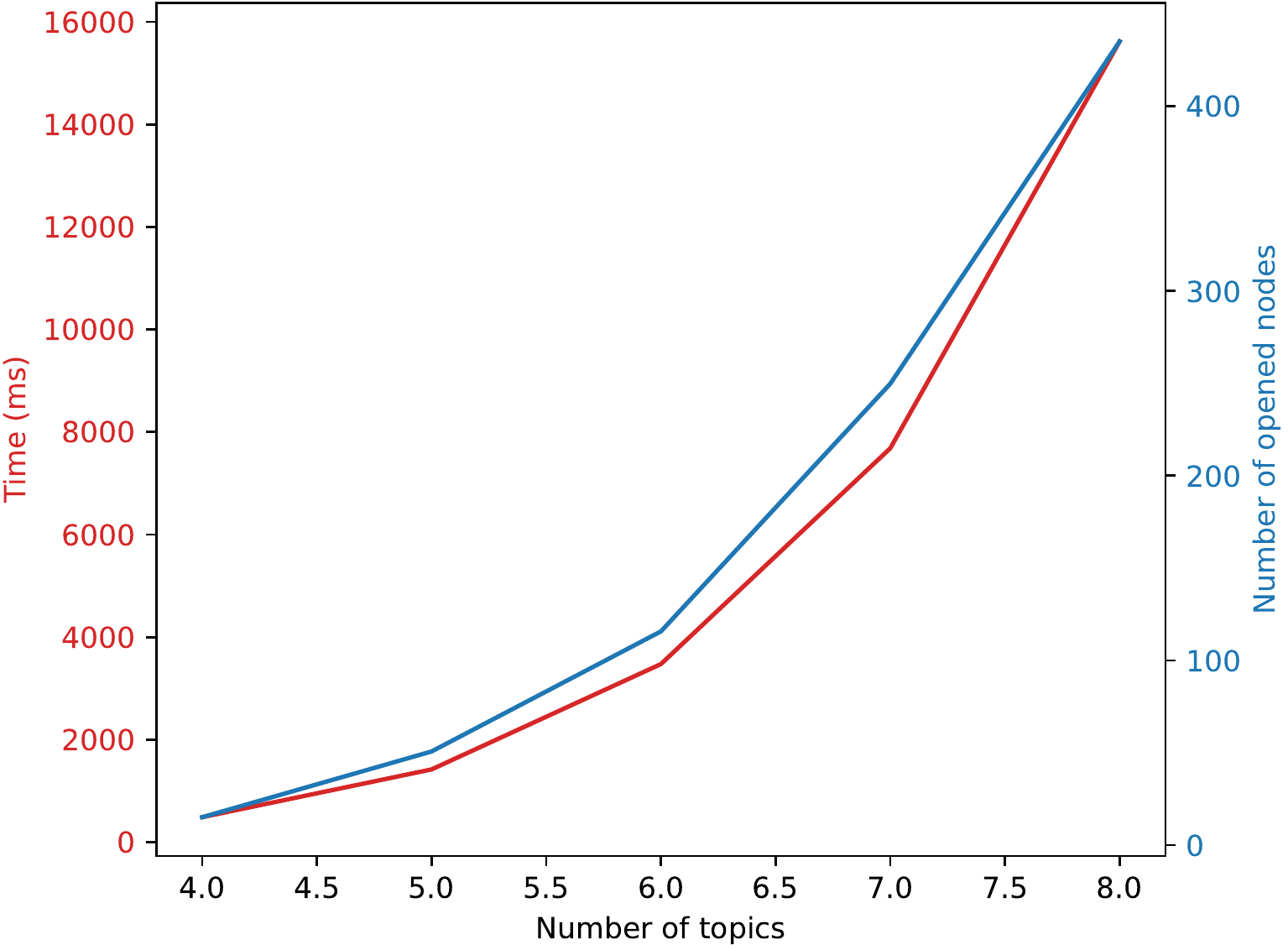}
\caption{(left) Losses over all the instances with respect to the number of topics. (right) Time and number of opened nodes with respect to the number of topics.} 
    \label{fig:nmflosses}
\end{figure}

Figure~\ref{fig:nmflosses} (left) shows the losses for the true reconstruction using the $W$ and $H$ matrices, 
and the best lost obtained by the algorithm. 
The true reconstruction loss is high because noises have been added to all instances.
Such results, up with the Figure~\ref{fig:nmfPriorUse}, shows that the algorithm is able to use prior knowledge, to find good decomposition, and to significantly handle the noise.

Finally, figure~\ref{fig:nmflosses} (right) shows the exponential behavior of the tree search, both in term of time or nodes.
Such a result implies that for large scale instances, users will have to use classical existing tools reducing the search time such as restarts, or custom searches.

This final experiments showed that the \bl{} framework up with the extended table constraints efficiently applied the prior knowledge to our learning models. 

\section{Conclusion}
This paper proposes to combine classical tools from combinatorial optimization and machine learning to tackle constrained machine learning problems.
The main advantage of such an approach is to reuse the large effort of combinatorial optimization in the context of efficient machine learning constraining. A simple yet generic framework branch and bound like algorithm named \bl{} is proposed to solve constrained learning problems involving combinatorial constraints.
Moreover, this paper introduces the extended table constraint, a constraint helping modelers to split the search space of the learning part.  The experimental parts shows that the proposed methodology can tackle problems that are known to be hard to solve otherwise.

\bibliography{main}

\bibliographystyle{plain}

\end{document}